%% file: manuscript.tex
\documentclass{ecai}  

\usepackage{graphicx}
\usepackage{latexsym}

\usepackage[toc,page]{appendix}
\input{preamble}

\begin{document}

\begin{frontmatter}

\title{Learning Logic Programs by Combining Programs}

\author
{\fnms{Andrew}~\snm{Cropper}\thanks{Corresponding Author. Email: andrew.cropper@cs.ox.ac.uk}}
\author
{\fnms{Céline}~\snm{Hocquette}}

\address{University of Oxford}

\input{00-abstract}

\end{frontmatter}

\input{01-intro}
\input{02-related}
\input{03-framework}
\input{04-imp}
\input{05-experiments}

\input{06-conclusions}

\section*{Code and Data}
The experimental code and data are available at \url{https://github.com/logic-and-learning-lab/ecai23-combo}.

\ack 
The authors are supported by the EPSRC fellowship (EP/V040340/1).
For open access, the authors have applied a CC BY public copyright licence to any author-accepted manuscript version arising from this submission.

\bibliography{ecai}

\begin{appendices}
 \input{A-background}
\input{A-encoding}
\input{correctness}

\input{D-domains}

\end{appendices}

\end{document}

%% file: preamble.tex
\newcommand{\name}{\textsc{combo}}
\newcommand{\popper}{\textsc{Popper}}
\newcommand{\metagol}{\textsc{Metagol}}
\newcommand{\tild}{\textsc{tilde}}
\newcommand{\ale}{\textsc{Aleph}}
\newcommand{\aspal}{\textsc{ASPAL}}
\newcommand{\progol}{\textsc{Progol}}
\newcommand{\ilasp}{\textsc{ilasp}}

\newcommand{\dcc}{\textsc{dcc}}

\usepackage[utf8]{inputenc}
\usepackage{booktabs}
\usepackage{amssymb}
\usepackage{listings}
\usepackage{inconsolata}
\usepackage{tikz}
\usepackage{pgfplots}
\usepackage{amsmath}
\usepackage{microtype}
\usepackage{algorithm}

\newtheorem{definition}{Definition}
\newtheorem{example}{Example}
\newtheorem{proposition}{Proposition}
\newtheorem{proof}{Proof}
\newtheorem{lemma}{Lemma}
\newtheorem{assumption}{Assumption}

\lstnewenvironment{myalgorithm}[1][] 
{
    \lstset{ 
        basicstyle=\ttfamily\small,
        showspaces=false,               
        showstringspaces=false,         
        mathescape=true,
        numbers=left,
        escapeinside={*}{*},
        columns=flexible,
        numbersep=3pt,        
        keywordstyle=\bfseries,
        keywords={,and, return, not, def, in, if, else, for, foreach, while, }
        numbers=left
        xleftmargin=1pt,
        #1 
    }
}
{}

%% file: 00-abstract.tex
\begin{abstract}
The goal of inductive logic programming is to induce a logic program (a set of logical rules) that generalises training examples.
Inducing programs with many rules and literals is a major challenge.
To tackle this challenge, we introduce an approach where we learn small \emph{non-separable} programs and combine them.
We implement our approach in a constraint-driven ILP system.
Our approach can learn optimal and recursive programs and perform predicate invention.
Our experiments on multiple domains, including game playing and program synthesis, show that our approach can drastically outperform existing approaches in terms of predictive accuracies and learning times, sometimes reducing learning times from over an hour to a few seconds.
\end{abstract}

%% file: 01-intro.tex
\section{Introduction}

The goal of inductive logic programming (ILP) \cite{mugg:ilp,ilpintro} is to induce a logic program (a set of logical rules) that generalises examples and background knowledge (BK).
The challenge is to efficiently search a large hypothesis space (the set of all programs) for a \emph{solution} (a program that correctly generalises the examples).

To tackle this challenge, divide-and-conquer approaches \cite{tilde} divide the examples into subsets and search for a program for each subset.
Separate-and-conquer approaches \cite{progol} search for a rule that covers (generalises) a subset of the examples, separate these examples, and then search for more rules to cover the remaining examples.
Both approaches can learn programs with many rules and literals. 
However, as they only learn from a subset of the examples,
they cannot perform predicate invention and struggle to learn recursive and optimal programs so tend to overfit.

To overcome these limitations, recent approaches \cite{ilasp,hexmil,popper,apperception} use \emph{meta-level} search \cite{ilpintro} to learn optimal and recursive programs.
However, most recent approaches struggle to learn programs with many rules in a program, many literals in a rule, or both.
For instance, \aspal{} \cite{aspal} precomputes every possible rule allowed in a program and uses an answer set solver to find a subset that covers the examples.
However, this now widely adopted precomputation approach \cite{ilasp,hexmil,difflog,prosynth} does not scale to rules with more than a few literals.
Likewise, many recent approaches struggle to learn programs with more than a few rules \cite{dilp,meta_abduce,popper,hopper,DBLP:conf/icml/GlanoisJFWZ0LH22}.

In this paper, our goal is to overcome the scalability limitations of recent approaches yet maintain the ability to learn recursive and optimal programs and perform predicate invention.
The key idea is to first learn small \emph{non-separable} programs that cover some of the examples and then search for a combination (a union) of these programs that covers all the examples.

Non-separable programs can be seen as building blocks for larger programs.
A program $h$ is \emph{separable} when (i) it has at least two rules, and (ii) no predicate symbol in the head of a rule in $h$ also appears in the body of a rule in $h$. 
A program is \emph{non-separable} when it is not separable.
For instance, consider the program $p_1$:
\[
    \begin{array}{l}
    p_1 = \left\{
    \begin{array}{l}
        \emph{happy(A) $\leftarrow$ rich(A)}\\
        \emph{happy(A) $\leftarrow$ friend(A,B), famous(B)}\\
        \emph{happy(A) $\leftarrow$ married(A,B), beautiful(B)}\\
    \end{array}
    \right\}
    \end{array}
\]
This program has three rules which can be learned separately and combined.
In other words, the union of the logical consequences of each rule is equivalent to the logical consequences of $p_1$.
Therefore, $p_1$ is separable.
By contrast, consider the program $p_2$:
\[
    \begin{array}{l}
    p_2 = \left\{
    \begin{array}{l}
        \emph{happy(A) $\leftarrow$ rich(A)}\\
        \emph{happy(A) $\leftarrow$ married(A,B), happy(B)}\\
    \end{array}
    \right\}
    \end{array}
\]
This program has two rules that cannot be learned separately because the second recursive rule depends on the first rule.
In other words, the union of the logical consequences of each rule is not equivalent to the consequences of $p_2$.
Therefore, $p_2$ is non-separable.

To explore this idea, we build on \emph{learning from failures} (LFF) \cite{popper}.
LFF frames the learning problem as a constraint satisfaction problem (CSP),  where each solution to the CSP represents a program.
The goal of a LFF learner is to accumulate constraints to restrict the hypothesis space.
For instance, \popper{}, a LFF learner, uses a \emph{generate}, \emph{test}, and \emph{constrain} loop to generate programs and test them on the examples.
If a program is not a solution, \popper{} builds constraints to explain why and uses these constraints to restrict future program generation.
We use LFF to explore our idea because it supports learning optimal and recursive programs from infinite domains. Moreover, it is easily adaptable because of its declarative constraint-driven approach.
We build on LFF by (i) only generating non-separable programs in the generate stage, and (ii) adding a combine stage to search for combinations of non-separable programs.

\subsubsection*{Motivating Example}
We illustrate our approach with a simple example. 
Suppose we want to learn a program that generalises the following positive ($E^+$) and negative ($E^-$) examples of lists of numbers:
\begin{center}
\begin{tabular}{l}
\emph{E$^+$ = \{f([1,3,5,7]), f([6,9,4,4]), f([21,22,23,24])\}}\\
\emph{E$^-$ = \{f([2,3,1]), f([9,3,2]), f([21,22,1])\}}
\end{tabular}
\end{center}

\noindent
For instance, we might want to learn a program such as:
\[
    \begin{array}{l}
    h_0 = \left\{
    \begin{array}{l}
    \emph{f(A) $\leftarrow$ head(A,7)}\\
    \emph{f(A) $\leftarrow$ head(A,4), tail(A,B), head(B,4)}\\
    \emph{f(A) $\leftarrow$ head(A,23), tail(A,B), head(B,24)}\\
    \emph{f(A) $\leftarrow$ tail(A,B), f(B)}
    \end{array}
    \right\}
    \end{array}
\]

\noindent
This program says that the relation \emph{f(A)} holds for a list $A$ if $A$ contains the sequence [7] or [4,4] or [23,24].
The last recursive rule is important because it allows the program to generalise to lists of arbitrary length.

To find a program that generalises the examples, we use a generate, test, \emph{combine}, and constrain loop.
In the generate stage, we generate non-separable programs of increasing size (with one literal, two literals, etc), such as:
\[
    \begin{array}{l}
    h_1 = \left\{
    \begin{array}{l}
        \emph{f(A) $\leftarrow$ head(A,1)}\\
    \end{array}
    \right\}\\
    h_2 = \left\{
    \begin{array}{l}
        \emph{f(A) $\leftarrow$ head(A,2)}\\
    \end{array}
    \right\}\\
    h_3 = \left\{
    \begin{array}{l}
        \emph{f(A) $\leftarrow$ head(A,3)}\\
    \end{array}
    \right\}
    \end{array}
\]

\noindent
In the test stage, we test programs on the examples.
If a program covers a negative example then we build a \emph{generalisation} constraint to prune more general programs from the hypothesis space, as they will also cover the negative example. 
For instance, as $h_2$ covers the first negative example (\emph{f([2,3,1])} we prune $h_2$ and its generalisations, such as $h_4$:
\[
    \begin{array}{l}
    h_4 = \left\{
    \begin{array}{l}
    \emph{f(A) $\leftarrow$ head(A,2)}\\
    \emph{f(A) $\leftarrow$ tail(A,B), f(B)}\\
    \end{array}
    \right\}
    \end{array}
\]
If a program covers no positive examples then we build a \emph{specialisation} constraint to prune more specific programs from the hypothesis space, as they will also not cover any positive examples.
For instance, as we do not have a positive example where the first element is 3 we prune $h_3$ and its specialisations, such as $h_5$:
\[
    \begin{array}{l}
    h_5 = \left\{
    \begin{array}{l}
    \emph{f(A) $\leftarrow$ head(A,3), tail(A,B), head(B,9)}
    \end{array}
    \right\}
    \end{array}
\]

\noindent
If a program covers \emph{at least one} positive and no negative examples we add it to a set of \emph{promising} programs.
For instance, when generating programs of size five, suppose we generate the recursive program:
\[
    \begin{array}{l}
    h_6 = \left\{
    \begin{array}{l}
        \emph{f(A) $\leftarrow$ head(A,7)}\\
        \emph{f(A) $\leftarrow$ tail(A,B), f(B)}\\
    \end{array}
    \right\}
    \end{array}
\]
As $h_6$ covers at least one positive example (\emph{f([1,3,5,7])})
and no negative examples we deem it a promising program.

In our novel combine stage, we then search for a combination of promising programs that covers all the positive examples and is minimal in size. 
We formulate this combinational problem as an answer set programming  (ASP) problem \cite{asp}, for which there are highly performance solvers, such as Clingo \cite{clingo}.
If we cannot find a combination, we go to the constrain stage where we use any discovered constraints to generate a new program.
If we find a combination, we deem it the best solution so far.
For instance, suppose that after considering programs of size seven we see the programs:
\[
    \begin{array}{l}
    h_7 = \left\{
    \begin{array}{l}
        \emph{f(A) $\leftarrow$ head(A,4), tail(A,B), head(B,4)}\\
        \emph{f(A) $\leftarrow$ tail(A,B), f(B)}\\
    \end{array}
    \right\}\\
    h_8 = \left\{
    \begin{array}{l}
    \emph{f(A) $\leftarrow$ head(A,23), tail(A,B), head(B,24)}\\
    \emph{f(A) $\leftarrow$ tail(A,B), f(B)}\\
    \end{array}
    \right\}
    \end{array}
\]
\noindent
Then the combination of $h_6 \cup h_7 \cup h_8$ is $h_0$, the solution we want to learn.
Crucially, we have learned a program with 4 rules and 13 literals by only generating programs with at most 2 rules and 7 literals. 
As the search complexity of ILP approaches is usually exponential in the size of the program to be learned, this reduction can substantially improve learning performance.

At this point, we have not proven that the combination is optimal in terms of program size.
In other words, we have not proven that there is no smaller solution.
Therefore, to learn an optimal solution, we continue to the constrain stage and add a constraint on the maximum program size (at most 12 literals) in future iterations.
This constraint prohibits (i) any program with more than 12 literals from being generated in the generate stage, and (ii) any combination of promising programs with more than 12 literals from being found in the combine stage.
We repeat this loop until we prove the optimality of a solution. 

\paragraph{Novelty, impact, and contributions.}
The main novelty of this paper is the idea of \emph{learning small non-separable programs that cover some of the examples and combining these programs to learn large programs with many rules and literals}.
We expand on this novelty in Section \ref{sec:related}.
The impact, which our experiments conclusively show on many diverse domains, is vastly improved learning performance, both in terms of predictive accuracies and learning times, sometimes reducing learning times from over one hour to a few seconds.
Moreover, as the idea connects many areas of AI , including program synthesis, constraint satisfaction, and logic programming, the idea should interest a broad audience.

Overall, we make the following contributions:

\begin{itemize}
    \item We introduce a generate, test, combine, and constrain ILP approach.
    \item We implement our idea in \name{}, a new system that learns optimal and recursive programs and supports predicate invention.
    We prove that \name{} always returns an optimal solution if one exists.
    \item We experimentally show on many diverse domains, including game playing and program synthesis, that our approach can substantially outperform other approaches, especially in terms of learning times. 
\end{itemize}

%% file: 02-related.tex
\section{Related Work}
\label{sec:related}

\paragraph{Rule mining.}
ILP is a form of rule mining.
A notable rule mining approach is AMIE+ \cite{DBLP:journals/vldb/GalarragaTHS15}.
Comparing \name{} with AMIE+ is difficult.
AMIE+ adopts an open-world assumption. 
By contrast, \name{} adopts the closed-world assumption. 
Moreover, \name{} can learn programs with relations of arity greater than two, which AMIE+ cannot, i.e. AMIE+ can only use unary and binary relations.
This difference is important as AMIE+ cannot be used on most of the datasets in our experiments. 
For instance, all the IGGP tasks \cite{iggp} and many of the program synthesis tasks use relations of arity greater than 2, such \emph{next\_cell/3}, \emph{true\_cell/3} and \emph{does\_jump/4} in the \emph{iggp-coins} task.
Likewise, many of the program synthesis tasks use \emph{append/3} or \emph{sum/3}. 
Finally, AMIE+ requires facts as input, which can be difficult to provide, especially when learning from infinite domains such as the program synthesis domain. 
By contrast, \name{} takes as input a definite program as BK.

\paragraph{Classic ILP.}
\tild{} \cite{tilde} is a divide-and-conquer approach.
\progol{} \cite{progol} is a separate-and-conquer approach that has inspired many other approaches \cite{xhail,atom,inspire}, notably \ale{} \cite{aleph}.
Although both approaches can learn programs with many rules and literals, they struggle to learn recursive and optimal programs and cannot perform predicate invention \cite{stahl:pi}.

\paragraph{Scalability.}
Scalability can be on many dimensions. 
For instance, many systems, such as QuickFOIL \cite{quickfoil}, focus on scaling to handle millions of training examples and background facts. 
Scaling to millions of examples is not a goal of this work (although we show that \name{} can handle hundreds of thousands of examples).
Instead, our goal is to scale to scale to large hypothesis spaces, which is difficult for many existing systems, notably rule selection approaches, outlined in the following paragraph.

\paragraph{Rule selection.}
Many systems formulate the ILP problem as a rule selection problem \cite{aspal,ilasp,dilp,hexmil}.
These approaches precompute every possible rule in the hypothesis space and then search for a subset that covers the examples, frequently using ASP to perform the search.
The major limitation of precomputation approaches is scalability in terms of (i) the size of rules, and (ii) the number of possible rules.
As these approaches precompute every possible rule, they cannot scale to rules with more than a few body literals because the number of rules is exponential in the number of body literals.
Similarly, as they perform a combinatorial search over every possible rule, they cannot scale to problems with many possible rules.
For instance, \textsc{Prosynth} \cite{prosynth} and \textsc{difflog} \cite{difflog} consider at most 1000 and 1267 candidate rules respectively.
However, our simplest experiment (\emph{trains1}) requires 31,860 candidate rules.
Our \emph{coins-goal} experiment requires approximately $10^{15}$ candidate rules, which is  infeasible for these approaches.
Moreover, our approach differs in many ways. 
We do not precompute every possible rule, which allows us to learn rules with many body literals.
In addition, we only search over promising programs (programs known to cover at least one positive and no negative example), which allows us to scale to problems with many possible rules.

\paragraph{LFF.}
Rather than precompute every possible rule, the key idea of \popper{} is to discover constraints from smaller programs (potentially with multiple rules) to rule out larger programs.
However, \popper{} struggles to learn large programs with many rules and many literals because it tries to generate a single program that covers all the examples.
We differ by (i) only generating non-separable programs in the generate step, and (ii) adding a combine step.
\dcc{} \cite{dcc} combines classical divide-and-conquer search with modern constraint-driven ILP.
\dcc{} learns a program for each example separately.
As these programs are likely to be overly specific, \dcc{} iteratively tries to learn more general programs.
To improve performance, \dcc{} reuses knowledge between iterations.
Our approach is completely different from \dcc{}.
\dcc{} tries to generate a single program, potentially separable, that covers all the examples and has no combine stage.
By contrast, \name{} never generates a separable program and searches for combinations of programs in the combine stage.
\textsc{hopper} \cite{hopper} extends \popper{} to learn higher-order programs.
Although our implementation only learns first-order programs, the approach should directly transfer to \textsc{hopper}.

%% file: 03-framework.tex
\section{Problem Setting}
\label{sec:setting}

We now describe our problem setting. 
We assume familiarity with logic programming \cite{lloyd:book} and ASP \cite{asp} but have included a summary in the appendix.

We use the LFF setting.
A \emph{hypothesis} is a set of definite clauses, i.e. 
we learn definite programs with the least Herbrand model semantics. 
We use the term \emph{program} interchangeably with the term hypothesis.
A \emph{hypothesis space} $\mathcal{H}$ is a set of hypotheses.
LFF uses \emph{hypothesis constraints} to restrict the hypothesis space.
Let $\mathcal{L}$ be a meta-language that defines hypotheses.
For instance, consider a meta-language formed of two literals \emph{h\_lit/3} and \emph{b\_lit/3} which represent \emph{head} and \emph{body} literals respectively.
With this language, we can denote the rule \emph{last(A,B) $\leftarrow$ tail(A,C), head(C,B)} as the set of literals \emph{\{h\_lit(0,last,(0,1)), b\_lit(0,tail,(0,2)), b\_lit(0,head,(2,1))\}}.
The first argument of each literal is the rule index, the second is the predicate symbol, and the third is the literal variables, where \emph{0} represents \emph{A}, \emph{1} represents \emph{B}, etc.
A \emph{hypothesis constraint} is a constraint (a headless rule) expressed in $\mathcal{L}$.
Let $C$ be a set of hypothesis constraints written in a language $\mathcal{L}$.
A hypothesis is \emph{consistent} with $C$ if when written in $\mathcal{L}$ it does not violate any constraint in $C$.
For instance, the rule \emph{last(A,B) $\leftarrow$ last(B,A)} violates the constraint \emph{$\leftarrow$ h\_lit(0,last,(0,1)), b\_lit(0,last,(1,0))}.
We denote as $\mathcal{H}_{C}$ the subset of the hypothesis space $\mathcal{H}$ which does not violate any constraint in $C$.

We define the LFF input:

\begin{definition}[\textbf{LFF input}]
\label{def:probin}
A \emph{LFF} input is a tuple $(E^+, E^-, B, \mathcal{H}, C)$ where $E^+$ and $E^-$ are sets of ground atoms denoting positive and negative examples respectively; $B$ is a definite program denoting background knowledge;
$\mathcal{H}$ is a hypothesis space, and $C$ is a set of hypothesis constraints.
\end{definition}

\noindent
We define a LFF solution:

\begin{definition}[\textbf{LFF solution}]
\label{def:solution}
Given an input tuple $(E^+, E^-, B, \mathcal{H}, C)$, a hypothesis $h \in \mathcal{H}_{C}$ is a \emph{solution} when $h$ is \emph{complete} ($\forall e \in E^+, \; B \cup h \models e$) and \emph{consistent} ($\forall e \in E^-, \; B \cup h \not\models e$).
\end{definition}

\noindent
If a hypothesis is not a solution then it is a \emph{failure}.
A hypothesis $h$ is \emph{incomplete} when $\exists e \in E^+, \; h \cup B \not \models e$; 
\emph{inconsistent} when $\exists e \in E^-, \; h \cup B \models e$;
 \emph{partially complete} when $\exists e \in E^+, \; h \cup B \models e$; 
 and \emph{totally incomplete} when $\forall e \in E^+, \; h \cup B \not \models e$.
 
Let $cost : \mathcal{H} \mapsto \mathbb{N}$ be an arbitrary cost function that measures the cost of a hypothesis.
We define an \emph{optimal} solution:

\begin{definition}[\textbf{Optimal solution}]
\label{def:opthyp}
Given an input tuple $(E^+, E^-, B, \mathcal{H}, C)$, a hypothesis $h \in \mathcal{H}_{C}$ is \emph{optimal} when (i) $h$ is a solution, and (ii) $\forall h' \in \mathcal{H}_{C}$, where $h'$ is a solution, $cost(h) \leq cost(h')$.
\end{definition}

\noindent
In this paper, our cost function is the number of literals in a hypothesis.

\paragraph{Constraints.}
The goal of a LFF learner is to learn hypothesis constraints from failed hypotheses.
Cropper and Morel \cite{popper} introduce hypothesis constraints based on subsumption \cite{plotkin:thesis}.
A clause $c_1$ \emph{subsumes} a clause $c_2$ ($c_1 \preceq c_2$) if and only if there exists a substitution $\theta$ such that $c_1\theta \subseteq c_2$.
A definite theory $t_1$ subsumes a definite theory $t_2$ ($t_1 \preceq t_2$) if and only if $\forall c_2 \in t_2, \exists c_1 \in t_1$ such that $c_1$ subsumes $c_2$.
A definite theory $t_1$ is a \emph{specialisation} of a definite theory $t_2$ if and only if $t_2 \preceq t_1$.
A definite theory $t_1$ is a \emph{generalisation} of a definite theory $t_2$ if and only if $t_1 \preceq t_2$.
A \emph{specialisation} constraint prunes specialisations of a hypothesis.
A \emph{generalisation} constraint prunes generalisations of a hypothesis.


%% file: 04-imp.tex
\section{Algorithm}
\label{sec:impl}

We now describe our \name{} algorithm.
To help explain our approach and delineate the novelty, we first describe \popper{}.

\paragraph{\popper{}.}

\noindent
\popper{} (Algorithm \ref{alg:popper}) solves the LFF problem.
\popper{} takes as input background knowledge (\emph{bk}), positive (\emph{pos}) and negative (\emph{neg}) examples, and an upper bound (\emph{max\_size}) on hypothesis sizes.
\popper{} uses  a generate, test, and constrain loop to find an optimal solution.
\popper{} starts with an ASP program $\mathcal{P}$ (hidden in the generate function).
The models of $\mathcal{P}$ correspond to hypotheses (definite programs).
In the generate stage (line 5), \popper{} uses Clingo \cite{clingo}, an ASP system, to search for a model of $\mathcal{P}$.
If there is no model, \popper{} increments the hypothesis size (line 7) and loops again.
If there is a model, \popper{} converts it to a hypothesis $h$.
In the test stage (line 9), \popper{} uses Prolog to test $h$ on the training examples.
We use Prolog because of its ability to handle lists and large, potentially infinite, domains.
If $h$ is a solution then \popper{} returns it.
If $h$ is a failure then, in the constrain stage (line 12), \popper{} builds hypothesis constraints (represented as ASP constraints) from the failure.
\popper{} adds these constraints to $\mathcal{P}$ to prune models, thus reducing the hypothesis space.
For instance, if $h$ is incomplete then \popper{} builds a specialisation constraint to prune its specialisations.
If $h$ is inconsistent then \popper{} builds a generalisation constraint to prune its generalisations.
\popper{} repeats this loop until it finds an optimal solution or there are no more hypotheses to test.

\begin{algorithm}[ht!]
{
\begin{myalgorithm}[]
def $\text{popper}$(bk, pos, neg, max_size):
  cons = {}
  size = 1
  while size $\leq$ max_size:
    h = generate(cons, size)
    if h == UNSAT:
      size += 1
      continue
    outcome = test(pos, neg, bk, h)
    if outcome == (COMPLETE, CONSISTENT)
      return h
    cons += constrain(h, outcome)
  return {}
\end{myalgorithm}
\caption{
\popper{}
}
\label{alg:popper}
}
\end{algorithm}

\subsection{\name{}}

\name{} (Algorithm \ref{alg:popp+}) builds on Algorithm \ref{alg:popper} but differs by (i) only building non-separable programs in the generate stage, and (ii) adding a \emph{combine} stage that tries to combine promising programs. 
We describe these novelties.

\begin{algorithm}[ht!]
{
\begin{myalgorithm}[]
def $\text{combo}$(bk, pos, neg, max_size):
  cons = {}
  promising = {}
  best_solution = {}
  size = 1
  while size $\leq$ max_size:
    h = generate_non_separable(cons, size)
    if h == UNSAT:
      size += 1
      continue
    outcome = test(pos, neg, bk, h)
    if outcome == (PARTIAL_COMPLETE, CONSISTENT):
      promising += h
      combine_outcome = combine(promising, max_size, bk,  neg)
      if combine_outcome != NO_SOLUTION:
        best_solution = combine_outcome
        max_size = size(best_solution)-1
    cons += constrain(h, outcome)
  return best_solution
\end{myalgorithm}
\caption{
\name{}
}
\label{alg:popp+}
}
\end{algorithm}

\subsubsection{Generate}
In the generate stage, \name{} only generates non-separable programs (line 7).
A program $h$ is \emph{separable} when (i) it has at least two rules, and (ii) no predicate symbol in the head of a rule in $h$ also appears in the body of a rule in $h$. 
A program is \emph{non-separable} when it is not separable.
For instance, \name{} cannot generate the following separable program:
\[
    \begin{array}{l}
    p_1 = \left\{
    \begin{array}{l}
        \emph{happy(A) $\leftarrow$ friend(A,B), famous(B)}\\
        \emph{happy(A) $\leftarrow$ married(A,B), beautiful(B)}\\
    \end{array}
    \right\}
    \end{array}
\]

\noindent
By only generating non-separable programs, we reduce the complexity of the generate stage.
Specifically, rather than search over every possible program, \name{} only searches over non-separable programs, a vastly smaller space that notably excludes all programs with multiple rules unless they are recursive or use predicate invention.
For instance, assume, for simplicity, that we have a problem with no recursion or predicate invention, that the rule space contains $m$ rules, and that we allow at most $k$ rules in a program.
Then in the generate stage, \popper{} searches over approximately $m^{k}$ programs.
By contrast, \name{} searches over only $m$ programs. 

To be clear, Algorithm \ref{alg:popp+} follows Algorithm \ref{alg:popper} and uses an ASP solver to search for a constraint-consistent (non-separable) program.
In other words, the \emph{generate\_non\_separable} function in Algorithm \ref{alg:popp+} is the same as the \emph{generate} function in Algorithm \ref{alg:popper} except it additionally tells the ASP solver to ignore separable programs using the encoding described in Section B.1 in the appendix.

\subsubsection{Test and Constrain}
If a program is partially complete (covers at least one positive example) and consistent, \name{} adds it to a set of promising programs (line 13).
If a program is inconsistent, \name{} builds a generalisation constraint to prune its generalisations from the hypothesis space, the same as \popper{}.
If a program is partially complete and inconsistent then, unlike \popper{}, \name{} does not build a specialisation constraint to prune its specialisations because we might want to specialise it.
For instance, consider learning a program to determine whether someone is happy.
Suppose \name{} generates the program:
\begin{center}
\begin{tabular}{l}
    \emph{happy(A) $\leftarrow$ rich(A)}
\end{tabular}
\end{center}
Suppose this program is partially complete and inconsistent.
Then we still might want to specialise this program to:
\begin{center}
\begin{tabular}{l}
    \emph{happy(A) $\leftarrow$ rich(A), tall(A)}
\end{tabular}
\end{center}
\noindent
\noindent
This program might now be partially complete and consistent, so is a promising program.
Therefore, \name{} only builds a specialisation constraint when a program is (i) totally incomplete, because none of its specialisations can be partially complete, or (ii) consistent, because there is no need to specialise a consistent program, as it could only cover fewer positive examples. 
We prove in the appendix that these constraints do not prune optimal hypotheses.

\subsubsection{Combine}
In the novel combine stage (line 14), \name{} searches for a combination (a union) of promising programs that covers the positive examples and is minimal in size.
We describe our combine algorithm in the next paragraph.
If we cannot find a combination, we go to the constrain stage where we use any discovered constraints to generate a new program (line 18).
If we find a combination, we deem it the best solution so far (line 16).
To provably learn an optimal solution, we update the maximum program size (line 17) which prohibits any program with more than \emph{max\_size} literals from being generated in the generate stage or any combination of promising programs with more than \emph{max\_size} literals from being considered in the combine stage.
We repeat this loop until we prove the optimality of a solution.

Algorithm \ref{alg:combine} shows the combine algorithm.
To find a combination of promising programs, we follow \aspal{} and formulate this combinational problem as an ASP problem.
The function \emph{build\_encoding} builds the encoding (line 5).
We briefly describe our encoding.
The appendix includes more details and an example encoding.
We give each positive example a unique ID.
For each rule in a promising program, we create a choice rule to indicate whether it should be in a solution.
For each promising program, we add facts about its example coverage and size.
We ask Clingo to find a model (a combination of rules) for the encoding (line 6) such that it (i) covers as many positive examples as possible, and (ii) is minimal in size.
If there is no model, we return the best solution so far; otherwise, we convert the model to a program (line 6).
Every combination program without recursion or predicate invention is guaranteed to be consistent (this result is an intermediate result in the proof of Theorem \ref{prop:correct} in the appendix).
However, if a combination program has recursion or predicate invention then it could be inconsistent.
In this case, we test the program on the negative examples to ensure consistency. 
If it is inconsistent, we add a constraint to the encoding (line 10) to eliminate this program and any generalistion of it from the combine encoding.
We then loop again.

\begin{algorithm}[ht!]
{
\begin{myalgorithm}[]
def $\text{combine}$(promising, max_size, bk, neg):
  cons = {}
  best_solution = NO_SOLUTION
  while True:
    encoding = build_encoding(promising, cons, max_size)
    h = call_clingo(encoding)
    if h == UNSAT:
        break
    if recursion_or_pi(h) and inconsistent(h, bk, neg):
        cons += build_con(h)
    else:
        best_solution = h
        break
  return best_solution
\end{myalgorithm}
\caption{
Combine
}
\label{alg:combine}
}
\end{algorithm}

A key advantage of our approach is that, whereas most rule selection approaches (Section \ref{sec:related}), including \aspal{}, assign a choice rule to every possible rule in the hypothesis space, we only do so to rules in promising programs. 
This difference is crucial to the performance of \name{} as it greatly reduces the complexity of the combinatorial problem.
For instance, let $m$ be the total number of possible rules and $n$ be the number of promising programs.
Then precomputation approaches search over 2$^{m}$ programs whereas \name{} searches over $2^{n}$ programs.
In practice $n$ is vastly smaller than $m$ so our combine stage is highly efficient.
For instance, in our simplest experiment (\emph{trains1}) there are 31,860 candidate rules so precomputation approaches search over $2^{31860}$ programs.
By contrast, \name{} finds an optimal solution in four seconds by only searching over 10 promising programs, i.e. over $2^{10}$ programs.

\paragraph{Correctness.} 
We prove the correctness of \name{}\footnote{
\name{} does not return every optimal solution.
To do so, we can revise Algorithm \ref{alg:combine} to ask the ASP solver to find and return all combinations at a certain size.
Algorithm \ref{alg:popp+} would then maintain a set of all current best programs, rather than a single program. 
}:

\begin{theorem}[\textbf{Correctness}] 
\label{prop:correct}
\name{} returns an optimal solution if one exists.
\end{theorem}
Due to space limitations, the proof is in the appendix.
At a high-level, to show this result, we show that (i) \name{} can generate and test every non-separable program, (ii) an optimal separable solution can be formed from a union (combination) of non-separable programs, and (iii) our constraints never prune optimal solutions.

%% file: 05-experiments.tex
\section{Experiments}
\label{sec:exp}
Our experiments aim to answer the question:
\begin{description}
\item[Q1] Can combining non-separable programs improve predictive accuracies and learning times?
\end{description}

\noindent
To answer \textbf{Q1}, we compare the performance of \name{} against \popper{}.
As \name{} builds on \popper{}, this comparison directly measures the impact of our new idea, i.e. it is the only difference between the systems.

To see whether \name{} is competitive against other approaches, our experiments aim to answer the question:

\begin{description}
\item[Q2] How does \name{} compare against other approaches?
\end{description}

\noindent
To answer \textbf{Q2} we also compare \name{} against \dcc{}, \ale{}, and \metagol{}  \cite{mugg:metagold}\footnote{
    We also tried to compare \name{} against rule selection approaches.
    However, precomputing every possible rule is infeasible for our datasets.
    The appendix contains more details.
}.
We use these systems because they can learn recursive definite programs.
\dcc{}, \popper{}, and \name{} use identical biases so the comparison between them is fair.
\ale{} uses a similar bias but has additional settings.
We have tried to make a fair comparison but there will likely be different settings that improve the performance of \ale{}.
The results for \metagol{} are in the appendix.




\subsubsection*{Methods}
We measure predictive accuracy and learning time given a maximum learning time of 60 minutes.
If a system does not terminate within the time limit, we take the best solution found by the system at that point.
We repeat all the experiments 10 times and calculate the mean and standard error\footnote{\url{https://en.wikipedia.org/wiki/Standard_error}}.
The error bars in the tables denote standard error.
We round learning times over one second to the nearest second because the differences are sufficiently large that finer precision is unnecessary. 
We use a 3.8 GHz 8-Core Intel Core i7 with 32GB of ram.
All the systems use a single CPU.



\subsubsection*{Domains}

We use the following domains.
The appendix contains more details, such as example solutions for each task and statistics about the problem sizes.

\textbf{Trains.}
The goal is to find a hypothesis that distinguishes eastbound and westbound trains \cite{michalski:trains}.

\textbf{Chess.} 
The task is to learn chess patterns in the king-rook-king (\emph{krk}) endgame \cite{celine:bottom}.
This dataset contains relations with arity greater than two, such as \emph{distance/3} and \emph{cell/4}.

\textbf{Zendo.}
Zendo is a multiplayer game where players must discover a secret rule by building structures. 
Zendo is a challenging game that has attracted much interest in cognitive science \cite{zendo}.

\textbf{IMDB.}
This real-world dataset \cite{mihalkova2007} contains relations between movies, actors, and directors.
This dataset is frequently used to evaluate rule learning systems \cite{alps}.
Note that this dataset has non-trivial numbers of training examples. 
For instance, the \emph{imdb3} task has 121,801 training examples.

\textbf{IGGP.}
The goal of \emph{inductive general game playing} \cite{iggp} (IGGP) is to induce rules to explain game traces from the general game playing competition \cite{ggp}.
This dataset is notoriously difficult for ILP systems: the currently best-performing system can only learn perfect solutions for 40\% of the tasks.
Moreover, although seemingly a toy problem, IGGP is representative of many real-world problems, such as inducing semantics of programming languages \cite{DBLP:conf/ilp/BarthaC19}. 
We use six games: \emph{minimal decay (md)}, \emph{rock - paper - scissors (rps)}, \emph{buttons}, \emph{attrition}, \emph{centipede}, and \emph{coins}.
These tasks all require learning rules with relations of arity more than two, which is impossible for some rule learning approaches \cite{DBLP:journals/vldb/GalarragaTHS15}.

\textbf{Graph problems.}
We use frequently used graph problems \cite{dilp,DBLP:conf/icml/GlanoisJFWZ0LH22}.
All of these tasks require the ability to learn recursive programs.





\textbf{Program synthesis.}
Inducing complex recursive programs is a difficult problem \cite{ilp20} and most ILP systems cannot learn recursive programs.
We use a program synthesis dataset \cite{popper} augmented with two new tasks \emph{contains} and \emph{reverse}. 
The motivating example in the introduction describes the task \emph{contains}.
These tasks all require the ability to learn recursive programs and to learn from non-factual data.

\subsection{Experimental Results}

\subsubsection*{Q1. Can combining non-separable programs improve predictive accuracies and learning times?}
Table \ref{tab:accs} shows that \name{} (i)  has equal or higher predictive accuracy than \popper{} on all the tasks, and (ii) can improve predictive accuracy on many tasks.
A McNemar's test confirms the significance of the difference at the p $<$ 0.01 level.
\name{} comprehensively outperforms \popper{} when learning programs with many rules and literals.
For instance, the solution for \emph{buttons} (included in the appendix) has 10 rules and 61 literals.
For this problem, \popper{} cannot learn a solution in an hour so has default accuracy.
By contrast, \name{} learns an accurate and optimal solution in 23s.

\begin{table}[t]
\centering
\begin{tabular}{@{}l|cccc@{}}
\textbf{Task} & \textbf{\name{}} & \textbf{\popper{}} & \textbf{\dcc{}} & \textbf{\ale{}}\\
\midrule
\emph{trains1} & \textbf{100 $\pm$ 0}  & \textbf{100 $\pm$ 0}  & \textbf{100 $\pm$ 0}  & \textbf{100 $\pm$ 0}  \\
\emph{trains2} & 98 $\pm$ 0 & 98 $\pm$ 0 & 98 $\pm$ 0 & \textbf{100 $\pm$ 0}  \\
\emph{trains3} & \textbf{100 $\pm$ 0}  & 79 $\pm$ 0 & \textbf{100 $\pm$ 0}  & \textbf{100 $\pm$ 0}  \\
\emph{trains4} & \textbf{100 $\pm$ 0}  & 32 $\pm$ 0 & \textbf{100 $\pm$ 0}  & \textbf{100 $\pm$ 0}  \\
\midrule
\emph{zendo1} & \textbf{97 $\pm$ 0}  & \textbf{97 $\pm$ 0}  & \textbf{97 $\pm$ 0}  & 90 $\pm$ 2 \\
\emph{zendo2} & \textbf{93 $\pm$ 2}  & 50 $\pm$ 0 & 81 $\pm$ 3 & \textbf{93 $\pm$ 3}  \\
\emph{zendo3} & \textbf{95 $\pm$ 2}  & 50 $\pm$ 0 & 78 $\pm$ 3 & \textbf{95 $\pm$ 2}  \\
\emph{zendo4} & \textbf{93 $\pm$ 1}  & 54 $\pm$ 4 & 88 $\pm$ 1 & 88 $\pm$ 1 \\
\midrule
\emph{imdb1} & \textbf{100 $\pm$ 0}  & \textbf{100 $\pm$ 0}  & \textbf{100 $\pm$ 0}  & \textbf{100 $\pm$ 0}  \\
\emph{imdb2} & \textbf{100 $\pm$ 0}  & \textbf{100 $\pm$ 0}  & \textbf{100 $\pm$ 0}  & 50 $\pm$ 0 \\
\emph{imdb3} & \textbf{100 $\pm$ 0}  & 50 $\pm$ 0 & \textbf{100 $\pm$ 0}  & 50 $\pm$ 0 \\
\midrule
\emph{krk1} & \textbf{98 $\pm$ 0}  & \textbf{98 $\pm$ 0}  & \textbf{98 $\pm$ 0}  & 97 $\pm$ 0 \\
\emph{krk2} & 79 $\pm$ 4 & 50 $\pm$ 0 & 54 $\pm$ 4 & \textbf{95 $\pm$ 0}  \\
\emph{krk3} & 54 $\pm$ 0 & 50 $\pm$ 0 & 50 $\pm$ 0 & \textbf{90 $\pm$ 4}  \\
\midrule
\emph{md} & \textbf{100 $\pm$ 0}  & 37 $\pm$ 13 & \textbf{100 $\pm$ 0}  & 94 $\pm$ 0 \\
\emph{buttons} & \textbf{100 $\pm$ 0}  & 19 $\pm$ 0 & \textbf{100 $\pm$ 0}  & 96 $\pm$ 0 \\
\emph{rps} & \textbf{100 $\pm$ 0}  & 18 $\pm$ 0 & \textbf{100 $\pm$ 0}  & \textbf{100 $\pm$ 0}  \\
\emph{coins} & \textbf{100 $\pm$ 0}  & 17 $\pm$ 0 & \textbf{100 $\pm$ 0}  & 17 $\pm$ 0 \\
\emph{buttons-g} & \textbf{100 $\pm$ 0}  & 50 $\pm$ 0 & 86 $\pm$ 1 & \textbf{100 $\pm$ 0}  \\
\emph{coins-g} & \textbf{100 $\pm$ 0}  & 50 $\pm$ 0 & 90 $\pm$ 6 & \textbf{100 $\pm$ 0}  \\
\emph{attrition} & \textbf{98 $\pm$ 0}  & 2 $\pm$ 0 & 2 $\pm$ 0 & \textbf{98 $\pm$ 0}  \\
\emph{centipede} & \textbf{100 $\pm$ 0}  & \textbf{100 $\pm$ 0}  & 81 $\pm$ 6 & \textbf{100 $\pm$ 0}  \\
\midrule
\emph{adj\_red} & \textbf{100 $\pm$ 0}  & \textbf{100 $\pm$ 0}  & \textbf{100 $\pm$ 0}  & 50 $\pm$ 0 \\
\emph{connected} & \textbf{98 $\pm$ 0}  & 81 $\pm$ 7 & 82 $\pm$ 7 & 51 $\pm$ 0 \\
\emph{cyclic} & \textbf{89 $\pm$ 3}  & 80 $\pm$ 7 & 85 $\pm$ 5 & 50 $\pm$ 0 \\
\emph{colouring} & \textbf{98 $\pm$ 1}  & \textbf{98 $\pm$ 1}  & \textbf{98 $\pm$ 1}  & 50 $\pm$ 0 \\
\emph{undirected} & \textbf{100 $\pm$ 0}  & \textbf{100 $\pm$ 0}  & \textbf{100 $\pm$ 0}  & 50 $\pm$ 0 \\
\emph{2children} & \textbf{100 $\pm$ 0}  & 99 $\pm$ 0 & \textbf{100 $\pm$ 0}  & 50 $\pm$ 0 \\
\midrule
\emph{dropk} & \textbf{100 $\pm$ 0}  & \textbf{100 $\pm$ 0}  & \textbf{100 $\pm$ 0}  & 55 $\pm$ 4 \\
\emph{droplast} & \textbf{100 $\pm$ 0}  & 95 $\pm$ 5 & \textbf{100 $\pm$ 0}  & 50 $\pm$ 0 \\
\emph{evens} & \textbf{100 $\pm$ 0}  & \textbf{100 $\pm$ 0}  & \textbf{100 $\pm$ 0}  & 50 $\pm$ 0 \\
\emph{finddup} & \textbf{99 $\pm$ 0}  & 98 $\pm$ 0 & \textbf{99 $\pm$ 0}  & 50 $\pm$ 0 \\
\emph{last} & \textbf{100 $\pm$ 0}  & \textbf{100 $\pm$ 0}  & \textbf{100 $\pm$ 0}  & 55 $\pm$ 4 \\
\emph{contains} & \textbf{100 $\pm$ 0}  & \textbf{100 $\pm$ 0}  & 99 $\pm$ 0 & 56 $\pm$ 2 \\
\emph{len} & \textbf{100 $\pm$ 0}  & \textbf{100 $\pm$ 0}  & \textbf{100 $\pm$ 0}  & 50 $\pm$ 0 \\
\emph{reverse} & \textbf{100 $\pm$ 0}  & 85 $\pm$ 7 & \textbf{100 $\pm$ 0}  & 50 $\pm$ 0 \\
\emph{sorted} & \textbf{100 $\pm$ 0}  & \textbf{100 $\pm$ 0}  & \textbf{100 $\pm$ 0}  & 74 $\pm$ 2 \\
\emph{sumlist} & \textbf{100 $\pm$ 0}  & \textbf{100 $\pm$ 0}  & \textbf{100 $\pm$ 0}  & 50 $\pm$ 0 \\

\end{tabular}
\caption{
Predictive accuracies.
}
\label{tab:accs}
\end{table}

\begin{table}[t]
\centering
\begin{tabular}{@{}l|cccc@{}}
\textbf{Task} & \textbf{\name{}} & \textbf{\popper{}} & \textbf{\dcc{}} & \textbf{\ale{}}\\
\midrule
\emph{trains1} & 4 $\pm$ 0 & 5 $\pm$ 0 & 8 $\pm$ 1 & \textbf{3 $\pm$ 1}  \\
\emph{trains2} & 4 $\pm$ 0 & 82 $\pm$ 25 & 10 $\pm$ 1 & \textbf{2 $\pm$ 0}  \\
\emph{trains3} & 18 $\pm$ 1 & \emph{timeout} & \emph{timeout} & \textbf{13 $\pm$ 3}  \\
\emph{trains4} & \textbf{16 $\pm$ 1}  & \emph{timeout} & \emph{timeout} & 136 $\pm$ 55 \\
\midrule
\emph{zendo1} & 3 $\pm$ 1 & 7 $\pm$ 1 & 7 $\pm$ 1 & \textbf{1 $\pm$ 0}  \\
\emph{zendo2} & 49 $\pm$ 5 & \emph{timeout} & 3256 $\pm$ 345 & \textbf{1 $\pm$ 0}  \\
\emph{zendo3} & 55 $\pm$ 6 & \emph{timeout} & \emph{timeout} & \textbf{1 $\pm$ 0}  \\
\emph{zendo4} & 53 $\pm$ 11 & 3243 $\pm$ 359 & 2939 $\pm$ 444 & \textbf{1 $\pm$ 0}  \\
\midrule
\emph{imdb1} & \textbf{2 $\pm$ 0}  & 3 $\pm$ 0 & 3 $\pm$ 0 & 142 $\pm$ 41 \\
\emph{imdb2} & \textbf{3 $\pm$ 0}  & 11 $\pm$ 1 & \textbf{3 $\pm$ 0}  & \emph{timeout} \\
\emph{imdb3} & \textbf{547 $\pm$ 46}  & 875 $\pm$ 166 & 910 $\pm$ 320 & \emph{timeout} \\
\midrule
\emph{krk1} & 28 $\pm$ 6 & 1358 $\pm$ 321 & 188 $\pm$ 53 & \textbf{3 $\pm$ 1}  \\
\emph{krk2} & 3459 $\pm$ 141 & \emph{timeout} & \emph{timeout} & \textbf{11 $\pm$ 4}  \\
\emph{krk3} & \emph{timeout} & \emph{timeout} & \emph{timeout} & \textbf{16 $\pm$ 3}  \\
\midrule
\emph{md} & 13 $\pm$ 1 & 3357 $\pm$ 196 & \emph{timeout} & \textbf{4 $\pm$ 0}  \\
\emph{buttons} & \textbf{23 $\pm$ 3}  & \emph{timeout} & \emph{timeout} & 99 $\pm$ 0 \\
\emph{rps} & 87 $\pm$ 15 & \emph{timeout} & \emph{timeout} & \textbf{20 $\pm$ 0}  \\
\emph{coins} & \textbf{490 $\pm$ 35}  & \emph{timeout} & \emph{timeout} & \emph{timeout} \\
\emph{buttons-g} & \textbf{3 $\pm$ 0}  & \emph{timeout} & \emph{timeout} & 86 $\pm$ 0 \\
\emph{coins-g} & 105 $\pm$ 6 & \emph{timeout} & \emph{timeout} & \textbf{9 $\pm$ 0}  \\
\emph{attrition} & \textbf{26 $\pm$ 1}  & \emph{timeout} & \emph{timeout} & 678 $\pm$ 25 \\
\emph{centipede} & \textbf{9 $\pm$ 0}  & 1102 $\pm$ 136 & 2104 $\pm$ 501 & 12 $\pm$ 0 \\
\midrule
\emph{adj\_red} & \textbf{2 $\pm$ 0}  & 5 $\pm$ 0 & 6 $\pm$ 0 & 479 $\pm$ 349 \\
\emph{connected} & \textbf{5 $\pm$ 1}  & 112 $\pm$ 71 & 735 $\pm$ 478 & 435 $\pm$ 353 \\
\emph{cyclic} & \textbf{35 $\pm$ 13}  & 1321 $\pm$ 525 & 1192 $\pm$ 456 & 1120 $\pm$ 541 \\
\emph{colouring} & \textbf{2 $\pm$ 0}  & 6 $\pm$ 0 & 5 $\pm$ 0 & 2373 $\pm$ 518 \\
\emph{undirected} & \textbf{2 $\pm$ 0}  & 6 $\pm$ 0 & 6 $\pm$ 0 & 227 $\pm$ 109 \\
\emph{2children} & \textbf{2 $\pm$ 0}  & 7 $\pm$ 0 & 6 $\pm$ 0 & 986 $\pm$ 405 \\
\midrule
\emph{dropk} & 7 $\pm$ 3 & 17 $\pm$ 2 & 14 $\pm$ 2 & \textbf{4 $\pm$ 1}  \\
\emph{droplast} & \textbf{3 $\pm$ 0}  & 372 $\pm$ 359 & 13 $\pm$ 1 & 763 $\pm$ 67 \\
\emph{evens} & 3 $\pm$ 0 & 29 $\pm$ 3 & 25 $\pm$ 2 & \textbf{2 $\pm$ 0}  \\
\emph{finddup} & 11 $\pm$ 5 & 136 $\pm$ 14 & 149 $\pm$ 7 & \textbf{0.8 $\pm$ 0}  \\
\emph{last} & \textbf{2 $\pm$ 0}  & 12 $\pm$ 1 & 11 $\pm$ 1 & \textbf{2 $\pm$ 0}  \\
\emph{contains} & \textbf{17 $\pm$ 0}  & 299 $\pm$ 52 & 158 $\pm$ 48 & 64 $\pm$ 5 \\
\emph{len} & 3 $\pm$ 0 & 52 $\pm$ 5 & 45 $\pm$ 2 & \textbf{2 $\pm$ 0}  \\
\emph{reverse} & 40 $\pm$ 5 & 1961 $\pm$ 401 & 1924 $\pm$ 300 & \textbf{3 $\pm$ 0}  \\
\emph{sorted} & 127 $\pm$ 78 & 111 $\pm$ 11 & 131 $\pm$ 10 & \textbf{1 $\pm$ 0}  \\
\emph{sumlist} & 4 $\pm$ 0 & 256 $\pm$ 27 & 221 $\pm$ 12 & \textbf{0 $\pm$ 0}  \\
\end{tabular}
\caption{
Learning times (seconds).
A \emph{timeout} entry means that the system did not terminate within 60 minutes.
}
\label{tab:times}
\end{table}


Table \ref{tab:times} shows that \name{} has lower learning times than \popper{} on all the tasks.
A paired t-test confirms the significance of the difference at the $p < 0.01$ level.
\name{} also outperforms \popper{} when learning small programs.
For instance, for \emph{centipede} both systems learn identical solutions with 2 rules and 8 literals.
However, whereas it takes \popper{} 1102s (18 minutes) to learn a solution, \name{} learns one in 9s, a 99\% reduction.
\name{} also outperforms \popper{} when learning recursive programs.
For instance, for \emph{reverse}, \popper{} needs 1961s (30 minutes) to learn a recursive program with 8 literals, whereas \name{} only needs 44s, a 98\% reduction.

To illustrate the efficiency of  \name{}, consider the \emph{coins-goal} task.
For this task, there are approximately $10^{15}$ possible rules\footnote{
For \emph{coins-goal}, 118 predicate symbols may appear in a rule, including symbols of arity 3 and 4.
Assuming at most 6 variables in a rule, there are around 1200 possible body literals.
Assuming at most 6 body literals, there are ${1200 \choose 6} \approx 10^{15}$ possible rules.
} and thus ${10^{15} \choose k}$ programs with $k$ rules -- which is why this task is infeasible for most rule selection approaches.
Despite this large hypothesis space, \name{} finds an optimal solution in under two minutes.
Moreover, the combine stage has a negligible contribution to the learning time.
For instance, in one trial that took \name{} 97s to learn a solution, only 0.07s was spent in the combine stage.



Overall, these results strongly suggest that the answer to \textbf{Q1} is yes: learning non-separable programs and combining them can drastically improve learning performance.

\subsubsection*{Q2. How does \name{} compare against other approaches?}
Table \ref{tab:accs} shows that \name{} has equal or higher accuracy than \dcc{} on all the tasks.
A McNemar's test confirms the significance of the differences at the p $<$ 0.01 level.
\name{} has notably higher accuracy on the \emph{iggp} and \emph{zendo} tasks.
Table \ref{tab:times} shows that \name{} has lower learning times than \dcc{} on all the tasks.
A paired t-test confirms the significance of the difference at the $p < 0.01$ level.
\name{} finds solutions for all but one task within the time limit.
By contrast, \dcc{} times out on 12 tasks.
For instance, for \emph{buttons-g} it takes \dcc{} over an hour to learn a solution with 86\% accuracy.
By contrast, \name{} finds a perfect solution in 3s, a 99\% reduction.
The appendix includes the learning output from \name{} for this task.

Table \ref{tab:accs} shows that \ale{} sometimes  has higher accuracy than \name{}, notably on the \emph{krk} tasks\footnote{
\name{} (and \popper{} and \dcc{}) struggles on this task because of an issue in the generate stage where the ASP solver struggles to find a model. 
We are unsure precisely why. 
To explore the issue, we ran Clingo with two threads, where each thread uses a different search heuristic. 
In this case, Clingo trivially finds a model, so we think that the default Clingo search heuristic just happens to struggle on the KRK tasks. 
To be clear, this issue is in the generate stage, not the combine stage.
}.
However, \name{} comfortably outperforms \ale{} on most tasks, especially graphs and synthesis, which require recursion.
\name{} notably outperforms \ale{} on the \emph{iggp} tasks which do not require recursion.
For instance, for \emph{coins}, \ale{} cannot find a solution in an hour.
By contrast, \name{} only needs 105s to learn a solution that has 100\% accuracy. 
The appendix includes a trace of the output from \name{} on this task. 

Table \ref{tab:times} shows that \ale{} sometimes has lower learning times than \name{}. 
The reason is that, unlike \name{}, \ale{} is not guaranteed to find an optimal solution, nor tries to, so terminates quicker.
For instance, for the \emph{rps} task, \ale{} is faster (20s) than \name{} (87s). 
However, \name{} takes only 2s to find the same solution as Aleph but takes 87s to prove that it is optimal. 

The results in the appendix show that \name{} also outperforms \metagol{} on all tasks.

Overall, these results strongly suggest that the answer to \textbf{Q2} is that \name{} compares favourably to existing approaches and can substantially improve learning performance, both in terms of predictive accuracies and learning times.

%% file: 06-conclusions.tex
\section{Conclusions and Limitations}
We have introduced an approach that learns small non-separable programs that cover some training examples and then tries to combine them to learn programs with many rules and literals.
We implemented this idea in \name{}, a new system that can learn optimal, recursive, and large programs and perform predicate invention.
We showed that \name{} always returns an optimal solution if one exists.
Our empirical results on many domains show that our approach can drastically improve predictive accuracies and reduce learning times compared to other approaches, sometimes reducing learning times from over 60 minutes to a few seconds.
In other words, \name{} can learn accurate solutions for  problems that other systems cannot.
These substantial improvements should directly help many application areas, such as drug design \cite{Kaalia16}, pathfinding \cite{forest-rubix}, and learning higher-order programs \cite{hopper}.

\subsubsection*{Limitations and Future Work}

\paragraph{Noise.}
In the combine stage, we search for a combination of programs that (i) covers all of the positive examples and none of the negative examples, and (ii) is minimal in size.
Our current approach is, therefore, intolerant to noisy/misclassified examples.
To address this limitation, we can relax condition (i) to instead find a combination that covers as many positive and as few negative examples.

\paragraph{Solvers.}
We formulate our combine stage as an ASP problem.
We could, however, use any constraint optimisation approach, such as formulating it as a MaxSAT \cite{maxsat} problem.
We therefore think this paper raises a challenge, especially to the constraint satisfaction and optimisation communities, of improving our approach by using different solvers and developing more efficient encodings.


%% file: A-background.tex
\section{Terminology}
\label{sec:bk}
\subsection{Logic Programming}
We assume familiarity with logic programming \cite{lloyd:book} but restate some key relevant notation. A \emph{variable} is a string of characters starting with an uppercase letter. A \emph{predicate} symbol is a string of characters starting with a lowercase letter. The \emph{arity} $n$ of a function or predicate symbol is the number of arguments it takes. An \emph{atom} is a tuple $p(t_1, ..., t_n)$, where $p$ is a predicate of arity $n$ and $t_1$, ..., $t_n$ are terms, either variables or constants. An atom is \emph{ground} if it contains no variables. A \emph{literal} is an atom or the negation of an atom. A \emph{clause} is a set of literals. A \emph{clausal theory} is a set of clauses. A \emph{constraint} is a clause without a positive literal. A \emph{definite} clause is a clause with exactly one positive literal. 
A \emph{hypothesis} is a set of definite clauses.
We use the term \emph{program} interchangeably with hypothesis, i.e. a \emph{hypothesis} is a \emph{program}.
A \emph{substitution} $\theta = \{v_1 / t_1, ..., v_n/t_n \}$ is the simultaneous replacement of each variable $v_i$ by its corresponding term $t_i$. 
A clause $c_1$ \emph{subsumes} a clause $c_2$ if and only if there exists a substitution $\theta$ such that $c_1 \theta \subseteq c_2$. 
A program $h_1$ subsumes a program $h_2$, denoted $h_1 \preceq h_2$, if and only if $\forall c_2 \in h_2, \exists c_1 \in h_1$ such that $c_1$ subsumes $c_2$. A program $h_1$ is a \emph{specialisation} of a program $h_2$ if and only if $h_2 \preceq h_1$. A program $h_1$ is a \emph{generalisation} of a program $h_2$ if and only if $h_1 \preceq h_2$

\subsection{Answer Set Programming}
We also assume familiarity with answer set programming \cite{asp} but restate some key relevant notation \cite{ilasp}.
A \emph{literal} can be either an atom $p$ or its \emph{default negation} $\text{not } p$ (often called \emph{negation by failure}). A normal rule is of the form $h \leftarrow b_1, ..., b_n, \text{not } c_1,... \text{not } c_m$. where $h$ is the \emph{head} of the rule, $b_1, ..., b_n, \text{not } c_1,... \text{not } c_m$ (collectively) is the \emph{body} of the rule, and all $h$, $b_i$, and $c_j$ are atoms. A \emph{constraint} is of the form $\leftarrow b_1, ..., b_n, \text{not } c_1,... \text{not } c_m.$ where the empty head means false. A \emph{choice rule} is an expression of the form $l\{h_1,...,h_m\}u \leftarrow b_1,...,b_n, \text{not } c_1,... \text{not } c_m$ where the head $l\{h_1,...,h_m\}u$ is called an \emph{aggregate}. In an aggregate, $l$ and $u$ are integers and $h_i$, for $1 \leq i \leq m$, are atoms. An \emph{answer set program} $P$ is a finite set of normal rules, constraints and choice rules. Given an answer set program $P$, the \emph{Herbrand base} of $P$, denoted
as ${HB}_P$, is the set of all ground (variable free) atoms that can be formed from the predicates and constants that appear in $P$. When $P$ includes only normal rules, a set $A \in {HB}_P$ is an \emph{answer set} of $P$ iff it is the minimal model of the  \emph{reduct} $P^A$, which is the program constructed from the grounding of $P$ by first removing any rule whose body contains a literal $\text{not } c_i$ where $c_i \in A$, and then removing any defaultly negated literals in the remaining rules. An answer set $A$ satisfies a ground constraint $\leftarrow b_1, ..., b_n, \text{not } c_1,... \text{not } c_m.$ if it is not the case that $\{b_1, ..., b_n\} \in A$ and $A \cap \{c_1, ..., c_m\} = \emptyset$.

%% file: A-encoding.tex
\section{ASP encoding}
\label{sec:encoding}
\subsection{Generate} \label{generate}
We describe the encoding of the \emph{generate} stage.
Our encoding is the same as \popper{} except we disallow multi-clause hypotheses, except when they are recursive or include predicate invention \cite{poppi}.
To do so, we add the following ASP program to the generate encoding, where \emph{invented/2} denotes there is an invented predicate symbol in a program.

\begin{verbatim}
:- clause(1), not pi_or_rec.

pi_or_rec :- pi.
pi_or_rec :- recursive.

recursive:-
    head_literal(C,P,A,_),
    body_literal(C,P,A,_).

pi:-
    invented(_,_).
\end{verbatim}


\subsection{Combine}
We describe the ASP encoding of the \emph{combine} stage. For each positive example, \name{} builds a unique example id. For instance, if the training set contains 6 positive examples, \name{} adds the following facts to the ASP encoding:
\begin{verbatim}
example(1).
example(2).
example(3).
example(4).
example(5).
example(6).
\end{verbatim}
For each promising program, \name{} builds a unique rule id for each rule in that program. It generates a fact describing the rule size. It then generates facts describing the examples covered by this promising program. For instance, assume a promising program has two rules of size 2 and 3 respectively. The first rule is given the id 15, and the second already had been used in a previous program and had already been given the id 3. Assume this program covers the examples with id $\{1,2,4\}$. Then, \name{} adds the following facts to the ASP encoding:
\begin{verbatim}
size(15,2).
covered(1) :- rule(15), rule(3).
covered(2) :- rule(15), rule(3).
covered(4) :- rule(15), rule(3).
\end{verbatim}
\name{} adds a choice rule which represents whether a rule is selected or not:
\begin{verbatim}
{rule(R)}:-size(R,_).
\end{verbatim}
\noindent
\name{} adds a soft constraint to cover the maximum number of positive examples:\\
\verb+:+$\sim$ \verb+ example(E), not covered(E). [1@2, (E,)]+\\
Finally, \name{} adds a soft constraint to prefer models/programs with the smallest size:\\
\verb+:+$\sim$ \verb+ rule(R),size(R,K). [K@1, (R,)]+\\
If a combination has been found in a previous iteration, \name{} bounds the program size to ensure newly selected combinations have smaller sizes. For instance, assume \name{} has found a combination of size 10 in a previous iteration. Then, programs with a size greater than 10 cannot be optimal. \name{} adds the following constraint to restrict the search accordingly:
\begin{verbatim}
:- #sum{K,R : rule(R), size(R,K)} >= 10.
\end{verbatim}
Finally, \name{} uses the ASP system Clingo \cite{clingo} to try to find a model of the encoding.

%% file: correctness.tex
\section{\name{} Correctness}
\label{sec:proof}
We show the correctness of \name{}.
In the rest of this section, we consider a LFF input tuple $(E^+, E^-, B, \mathcal{H}, C)$. We assume that an optimal solution always exists:
\begin{assumption}[\textbf{Existence of an optimal solution}] \label{existencesol}Given a LFF input tuple $(E^+, E^-, B, \mathcal{H}, C)$, we assume there exists a solution $h \in \mathcal{H}_{C}$.
\end{assumption}
\noindent
We follow LFF \cite{popper} and assume the BK does not depend on any hypothesis:
\begin{assumption}[\textbf{BK independence}] \label{independentbk}We assume that no predicate symbol in the body of a rule in the BK appears in the head of a rule in a hypothesis. 
\end{assumption}
\noindent
For instance, given the following BK we disallow learning hypotheses for the target relations \emph{famous} or \emph{friend}:
\begin{center}
\begin{tabular}{l}
\emph{happy(A) $\leftarrow$ friend(A,B), famous(B)}\\
\end{tabular}
\end{center}

\noindent
As with \popper{} \cite{popper}, \name{} is only correct when the hypothesis space only contains decidable programs (such as Datalog programs), i.e. when every program is guaranteed to terminate:
\begin{assumption}[\textbf{Decidable programs}]\label{decidable}
We assume the hypothesis space only contains decidable programs.
\end{assumption}
\noindent
When the hypothesis space contains arbitrary definite programs, the results do not hold because, due to their Turing completeness, checking entailment of an arbitrary definite program is semi-decidable \cite{tarnlund:hornclause}, i.e. testing a program might never terminate.

We show correctness in three steps. 
We first assume that there is no constrain stage, i.e. \name{} does not apply learned constraints.
With this assumption, we show that the generate, test, and combine stages return an optimal solution. 
We then show that the constrain stage is optimally sound in that it never prunes optimal solutions from the hypothesis space.
We finally use these two results to prove the correctness of \name{} i.e. that \name{} returns an optimal solution.

\subsection{Generate, Test, and Combine}
We define a separable program:

\begin{definition}
[\textbf{Separable program}]
A program $h$ is \emph{separable} when (i) it has at least two rules, and (ii) no predicate symbol in the head of a rule in $h$ also appears in the body of a rule in $h$.
\end{definition}

\noindent
In other words, a separable program cannot have recursive rules or invented predicate symbols.

\begin{example}[\textbf{Separable program}]%
The following program is separable because it has two rules and the head predicate symbol \emph{zendo2} does not appear in the body of a rule:
\begin{center}
\begin{tabular}{l}
\emph{zendo2(A) $\leftarrow$ piece(A,B),piece(A,D),piece(A,C),}\\
\hspace{53pt} \emph{green(D),red(B),blue(C)}\\
\emph{zendo2(A) $\leftarrow$ piece(A,D),piece(A,B),coord1(B,C),}\\
\hspace{53pt} \emph{green(D),lhs(B),coord1(D,C)}\\
\end{tabular}
\end{center}
\end{example}

\noindent
A program is \emph{non-separable} when it is not separable.
\begin{example}[\textbf{Non-separable program}]
The following program is non-separable because it has only one rule:
\begin{center}
\begin{tabular}{l}
\emph{zendo1(A) $\leftarrow$ piece(A,C),size(C,B),blue(C),}\\
\hspace{53pt} \emph{small(B),contact(C,D),red(D)}\\
\end{tabular}
\end{center}

\noindent
The following program is non-separable because the predicate symbol \emph{sorted} appears in the head of the first rule and the body of the second:
\begin{center}
\begin{tabular}{l}
\emph{sorted(A) $\leftarrow$ tail(A,B),empty(B)}\\
\emph{sorted(A) $\leftarrow$ tail(A,D),head(A,B),head(D,C),}\\
\hspace{50pt} \emph{geq(C,B),sorted(D)}\\
\end{tabular}
\end{center}

\noindent
The following program is non-separable because the predicate symbol \emph{contains} appears in the head of the first and second rules and the body of the third:
\begin{center}
\begin{tabular}{l}
\emph{contains(A) $\leftarrow$ head(A,B),c6(B)}\\
\emph{contains(A) $\leftarrow$ head(A,B),c9(B)}\\
\emph{contains(A) $\leftarrow$ tail(A,B),contains(B)}\\
\end{tabular}
\end{center}

\noindent
The following program is non-separable because the invented predicate symbol \emph{inv} appears in the body of the first rule and the head of the second and third rules:
\begin{center}
\begin{tabular}{l}
\emph{grandparent(A,B) $\leftarrow$ inv(A,C),inv(C,B)}\\
\emph{inv(A,B) $\leftarrow$ mother(A,B)}\\
\emph{inv(A,B) $\leftarrow$ father(A,B)}\\
\end{tabular}
\end{center}


\end{example}

\noindent
We show that \name{} can generate and test every non-separable program:
\begin{lemma}
\label{lem_generate}
\name{} can generate and test every non-separable program.
\end{lemma}
\begin{proof}
Cropper and Morel \cite{popper} show (Proposition 6) that the generate encoding of \popper{} has a model for every program in the hypothesis space.
Our generate encoding (Section \ref{generate}) generates all non-separable programs, a subset of all programs.
By Assumption \ref{decidable}, each program is guaranteed to terminate in the test stage.
Therefore \name{} can generate and test every non-separable program.
\end{proof}


\noindent
Lemma \ref{lem_generate} shows that \name{} can generate and test every non-separable program. 
However, the generate stage cannot generate separable programs.
To learn a separable program, \name{} uses the combine stage to combine non-separable programs.
To show the correctness of this combine stage, we first show that any separable program can be formed from the union of non-separable programs:
\begin{lemma}
\label{lem_separable}
Let $h$ be a separable program.
Then $h = \bigcup_{i=1}^n h_i$ where each $h_i$ is a non-separable program.


\end{lemma}
\begin{proof}
By induction on the number of rules $n$ in $h$. 
If $n=2$, then $h= \{r_1\} \cup \{r_2\}$ where $r_1$ and $r_2$ are the rules in $h$ and therefore are non-separable programs. 
For the inductive case, assume the claim holds for a program with $n$ rules.
We show the claim holds for a program $h$ with $n+1$ rules.
Let $r$ be the last rule of $h$.
By the induction hypothesis, $h \setminus \{r\} = \bigcup_{i=1}^n h_i$ where each $h_i$ is a non-separable program.
Therefore, $h = \{r\} \cup \bigcup_{i=1}^n h_i $ where $\{r\}$ is a non-separable program.
\end{proof}

\noindent
We define a \emph{promising} program:

\begin{definition}[\textbf{Promising program}]
 A program $h \in \mathcal{H}_{C}$ is \emph{promising} when it is partially complete ($\exists e \in E^+, \; B \cup h \models e$) and consistent ($\forall e \in E^-, \; B \cup h \not\models e$).
\end{definition}

\noindent
We show that any optimal solution is the union of non-separable promising programs:
\begin{lemma}\label{lem_opti_separable}
Let $I = (E^+, E^-, B, \mathcal{H}, C)$ be a LFF input, $h_o \in \mathcal{H}_{C}$ be an optimal solution for I, and $S$ be the set of all non-separable promising programs for $I$.
Then $h_o = \bigcup_{i=1}^n h_i$ where each $h_i$ is in $S$.
\end{lemma}

\begin{proof}
An optimal solution $h_o$ is either (a) non-separable, or (b) separable. 
For case (a), if $h_o$ is non-separable and 
a solution for $I$ then $h_o$ is promising program for $I$ and is thus in $S$. 
For case (b), if $h_o$ is separable then, by Lemma \ref{lem_separable}, $h_o = \bigcup_{i=1}^n h_i$ where each $h_i$ is a non-separable program.
For contradiction, assume some $h_i$ is not a promising program, which implies that $h_i$ is either (i) inconsistent, or (ii) totally incomplete. 
For case (i), if $h_i$ is inconsistent then $h_o$ is inconsistent and is not a solution so case (i) cannot hold.
For case (ii), let $h' = h_o \setminus h_i$. 
Since no predicate symbol in the body of $h'$ appears in the head of $h_i$ and, by Assumption \ref{independentbk}, no predicate symbol in the body of a rule in the BK appears in the head of a rule in $h_i$ then $h_i$ does not affect the logical consequences of $h'$.
Since $h_o$ is complete and $h_i$ is totally incomplete then $h'$ is complete.
Since $h_o$ is consistent then $h'$ is consistent. 
Therefore $h'$ is a solution. 
Since \emph{size($h'$) $<$ size($h_o$)} then $h_o$ cannot be optimal so case (ii) cannot hold.
These two cases are exhaustive so the assumption cannot hold and each $h_i$ is in $S$. 
Cases (a) and (b) are exhaustive so the proof is complete.
\end{proof}

\begin{example}[\textbf{Optimal separable solution}]
Consider the positive ($E^+$) and negative ($E^-$) examples:
\begin{center}
\begin{tabular}{l}
\emph{E$^+$ = \{f([1,3,5,7]), f([0,4,4]), f([1,2,4])\}}\\
\emph{E$^-$ = \{f([5,6,7]), f([3,4,2])\}}
\end{tabular}
\end{center}

\noindent
Also consider the optimal solution:
\[
    \begin{array}{l}
    h = \left\{
    \begin{array}{l}
\emph{f(A) $\leftarrow$ head(A,0)}\\
\emph{f(A) $\leftarrow$ head(A,1)}\\
    \end{array}
    \right\}
    \end{array}
\]
This solution can be formed from the union of two non-separable promising programs $h_1$ and $h_2$:
\[
    \begin{array}{l}
    h_1 = \left\{
    \begin{array}{l}
\emph{f(A) $\leftarrow$ head(A,0)}\\
    \end{array}
    \right\}\\
    h_2 = \left\{
    \begin{array}{l}
\emph{f(A) $\leftarrow$ head(A,1)}\\
    \end{array}
    \right\}
    \end{array}
\]
We show why Lemma \ref{lem_opti_separable} only holds for optimal solutions. 
Consider the hypothesis:
\[
    \begin{array}{l}
    h' = \left\{
    \begin{array}{l}
\emph{f(A) $\leftarrow$ head(A,0)}\\
\emph{f(A) $\leftarrow$ head(A,1)}\\
\emph{f(A) $\leftarrow$ head(A,2)}\\
    \end{array}
    \right\}
    \end{array}
\]
The hypothesis $h'$ is a solution for $E^+$ and $E^-$ but is not optimal. 
It can be formed from the union of the non-separable promising programs $h_1$ and $h_2$ and the non-separable program $h_3$:
\[
    \begin{array}{l}
    h_3 = \left\{
    \begin{array}{l}
\emph{f(A) $\leftarrow$ head(A,2)}\\
    \end{array}
    \right\}
    \end{array}
\]
However, $h_3$ is not promising because it does not cover any positive example.
\end{example}
\noindent


\noindent
We show that the combine stage returns an optimal solution:

\begin{lemma}
\label{lem_combine}
Let $I = (E^+, E^-, B, \mathcal{H}, C)$ be a LFF input and $S$ be the set of all non-separable promising programs for $I$. 
Then the combine stage returns an optimal solution for $I$.
\end{lemma}
\begin{proof}
For contradiction, assume the opposite, which implies that the combine stage returns a hypothesis $h$ that is (a) not a solution for $I$, or (b) a non-optimal solution for $I$.
For case (a), assume $h$ is not a solution. 
By Assumption \ref{existencesol} an optimal solution $h_o$ exists. From Lemma \ref{lem_opti_separable}, $h_o = \bigcup_{i=1}^n h_i$ where each $h_i \in S$. Since $h_o$ is a solution for $I$, $h_o$ is a model for the ASP encoding. Moreover, there are finitely many programs in $S$.
Since the ASP solver finds a model that covers as many positive examples as possible then $h$ is complete. 
As an output of the combine stage, $h$ is consistent. 
Therefore, $h$ is a solution so case (a) cannot hold.
For case (b), assume $h$ is a non-optimal solution. By Assumption \ref{existencesol} an optimal solution $h_o$ exists.
From Lemma \ref{lem_opti_separable}, $h_o = \bigcup_{i=1}^n h_i$ where each $h_i \in S$. 
Since $h_o$ is a solution for $I$ then it is a model for the ASP encoding.
The ASP program can reason about the size of combinations as the sum of the size of the rules in non-separable programs selected.
Moreover, there are finitely many programs in $S$.
Therefore, the ASP solver finds a model that is minimal in size and $h$ must be an optimal solution so case (b) cannot hold.
These cases are exhaustive so the assumption cannot hold and the combine stage returns an optimal solution.
\end{proof}

\noindent
We show that \name{} (with only the generate, test, and combine stages) returns a solution:
\begin{lemma}
\label{lem_gen_one_opt}
\name{} returns a solution.
\end{lemma}
\begin{proof}
Lemma \ref{lem_generate} shows that \name{} can generate and test every non-separable program, of which there are finitely many.
Therefore, \name{} can generate and test every promising program.
Lemma \ref{lem_combine} shows that given a set of promising programs, the combine stage returns a solution.
Therefore \name{} returns a solution.
\end{proof}
\noindent
We show that \name{} returns an optimal solution.
\begin{proposition}
\label{prop1}
    \name{} returns an optimal solution.
\end{proposition}
\begin{proof}
Follows from Lemma \ref{lem_gen_one_opt} and that \name{} uses iterative deepening over the program size to guarantee optimally.
\end{proof}




\subsection{Constrain}
In the previous section, we showed that \name{} finds an optimal solution when it does not apply any constraints, i.e. without the constrain stage.
In this section, we show that the constrain stage never prunes optimal solutions from the hypothesis space.
\name{} only builds constraints from non-separable programs that are (a) inconsistent, (b) consistent, or (c) totally incomplete.
We show a result for case (a):

\begin{lemma}
Let $h_1$ and $h_2$ be non-separable programs, $h_2$ be a generalisation of $h_1$ ($h_2 \prec h_1$), and $h_1$ be inconsistent.
Then $h_2$ is not in a solution.
\label{generalisation}
\end{lemma}
\begin{proof}
Since $h_1$ is inconsistent and $h_2 \prec h_1$ then $h_2$ is inconsistent and cannot be in a solution.
\end{proof}

\noindent
To show a result for case (b), we first show an intermediate result:

\begin{lemma}
\label{consistent}
Let $h_1$ and $h_2$ be non-separable consistent programs and $h = h_1 \cup h_2$ be a separable hypothesis. 
Then $h$ is consistent.
\end{lemma}
\begin{proof}
Since $h$ is separable then no predicate symbol in the head of a rule in $h_1$ (respectively $h_2$) appears in the body of a rule in $h_2$ (respectively $h_1$) or vice-versa.
In addition, by Assumption \ref{independentbk}, no predicate symbol in the body of a rule in the BK appears in the head of a rule in $h$.
Therefore $h_1$ does not affect the logical consequences of $h_2$ and vice-versa.
Therefore as $h_1$ and $h_2$ are both consistent then $h$ is consistent.
\end{proof}

\noindent
We use Lemma \ref{consistent} to show a result for case (b):

\begin{lemma}
\label{lem:spec_consistent}
Let $h_1$ and $h_2$ be non-separable programs, $h_2$ be a specialisation of $h_1$ ($h_1 \prec h_2$), and $h_1$ be consistent.
Then $h_2$ is not an optimal solution and is not in an optimal separable solution.
\end{lemma}
\begin{proof}
Assume the opposite, i.e. $h_2$ is (i) an optimal solution or (ii) in an optimal separable solution. For case (i), assume $h_2$ is an optimal solution. Since $h_2$ is complete and $h_2$ specialises $h_1$ then $h_1$ is complete. Since $h_1$ is complete and consistent then it is a solution. Since $h_2$ specialises $h_1$ then $size(h_1) < size(h_2)$.
Therefore, $size(h_1) < size(h_2)$ so $h_2$ is not an optimal solution and case (i) cannot hold.
For case (ii), assume $h_2$ is in an optimal separable solution $h_o$.
Then $h_o = h_2 \cup h_3$ where $h_3 \neq \emptyset$.
Let $h' = h_1 \cup h_3$. 
Since $h_o$ is complete and $h_2$ specialises $h_1$ then $h'$ is complete.
Since $h_o$ is consistent then $h_3$ is consistent. Since $h_o$ is separable and $h_2$ specialises $h_1$ then $h'$ is separable.
Since $h_1$ and $h_3$ are consistent and $h'$ is separable then Lemma \ref{consistent} shows that $h'$ is consistent.
As $h'$ is complete and consistent it is a solution.
Since $h_2$ specialises $h_1$ then $size(h_1) < size(h_2)$.
Therefore, $size(h') < size(h_o)$ so $h_o$ is not an optimal solution and case (ii) cannot hold.
These cases are exhaustive so the assumption cannot hold.
\end{proof}
\begin{example}
[\textbf{Specialisations of consistent programs}] 
Consider the hypothesis:
\[
    \begin{array}{l}
    h_1 = \left\{
    \begin{array}{l}
    \emph{f(A,B) $\leftarrow$ head(A,B)}
    \end{array}
    \right\}
    \end{array}
\]
If $h_1$ is consistent, we can prune its specialisations such as the following $h_2$ as they are not optimal solutions and cannot appear in an optimal separable solution:
\[
    \begin{array}{l}
    h_2 = \left\{
    \begin{array}{l}
    \emph{f(A,B) $\leftarrow$ head(A,B), even(B)}
    \end{array}
    \right\}
    \end{array}
\]

\noindent
It is important to note that case (b) does not prohibit specialisations of $h_1$ from being in multi-clause non-separable programs such as:
\[
    \begin{array}{l}
    h_3 = \left\{
    \begin{array}{l}
    \emph{f(A,B) $\leftarrow$ head(A,B),odd(B)}\\
    \emph{f(A,B) $\leftarrow$ tail(A,C), f(C,B)}\\
    \end{array}
    \right\}
    \end{array}
\]
\end{example}

\noindent
We show a result which corresponds to case (c):
\begin{lemma}
\label{lem_spec_total_complete}
Let $h_1$ and $h_2$ be non-separable programs, $h_2$ be a specialisation of $h_1$ ($h_1 \prec h_2$), and $h_1$ be totally incomplete.
 Then $h_2$ is not an optimal solution or in an optimal separable solution.
\end{lemma}
\begin{proof}
Assume the opposite, i.e. $h_2$ is (i) an optimal solution or (ii) in an optimal separable solution. For case (i), assume $h_2$ is an optimal solution. Since $h_1$ is totally incomplete and $h_2$ specialises $h_1$ then $h_2$ is totally incomplete and therefore cannot be a solution and case (i) cannot hold.
For case (ii), assume $h_2$ is in an optimal separable solution $h_o$. Then $h_o = h_2 \cup h_3$ where $h_3 \neq \emptyset$.
Since $h_1$ is totally incomplete and $h_2$ specialises $h_1$ then $h_2$ is totally incomplete.
Since $h_o$ is complete and separable and $h_2$ is totally incomplete then $h_3$ is complete.
Since $h_o$ is consistent then $h_3$ is consistent.
Since $h_3$ is complete and consistent it is a solution.
Since $h_3$ is a subset of $h_o$ then $size(h_3) < size(h_o)$.
Since $h_3$ is a solution and is smaller than $h_o$ then $h_o$ cannot be optimal and case (ii) cannot hold.
These cases are exhaustive so the assumption cannot hold.
\end{proof}

\begin{example}[
\textbf{Specialisations of totally incomplete programs}]
    Consider the hypothesis:
\[
    \begin{array}{l}
    h_1 = \left\{
    \begin{array}{l}
    \emph{f(A,B) $\leftarrow$ head(A,B)}
    \end{array}
    \right\}
    \end{array}
\]
If $h_1$ is totally incomplete, we can prune its specialisations, such as the following $h_2$, as they are not optimal solutions and cannot appear in an optimal separable solution.
\[
    \begin{array}{l}
    h_2 = \left\{
    \begin{array}{l}
    \emph{f(A,B) $\leftarrow$ head(A,B), even(B)}
    \end{array}
    \right\}
    \end{array}
\]

\noindent
It is important to note that case (c) does not prohibit specialisations of $h_1$ from being in multi-clause non-separable programs, such as:
\[
    \begin{array}{l}
    h_3 = \left\{
    \begin{array}{l}
    \emph{f(A,B) $\leftarrow$ head(A,B), odd(B)}\\
    \emph{f(A,B) $\leftarrow$ tail(A,C), f(C,B)}\\
    \end{array}
    \right\}
    \end{array}
\]
\end{example}
\noindent
While constraints inferred in case (a) never prune solutions, constraints inferred in cases (b) and (c) never prune \emph{optimal} solutions from the hypothesis space.
We show that the constrain stage never prunes optimal solutions:

\begin{proposition}
\label{prop_opt_sound}
The constrain stage of \name{} never prunes optimal solutions.
\end{proposition}
\begin{proof}
\name{} builds constraints from non-separable programs that are (i) inconsistent, (ii) consistent, or (iii) totally incomplete.
For case (i), if a program is inconsistent, \name{} prunes its generalisations which by Lemma \ref{generalisation} cannot be in a solution.
For case (ii), if a program is consistent, \name{} prunes its specialisations which by Lemma \ref{lem:spec_consistent} cannot be an optimal solution or in an optimal separable solution.
For case (iii), if a program is totally incomplete, \name{} prunes its specialisations which by Lemma \ref{lem_spec_total_complete} cannot be an optimal solution or in an optimal separable solution.
Therefore, \name{} never prunes optimal solutions.
\end{proof}

\subsection{\name{} Correctness}

\noindent
We finally show the correctness of the full \name{} generate, test, combine, and constrain algorithm:

\begin{theorem}
\name{} returns an optimal solution.
\end{theorem}
\begin{proof}
Follows from Propositions \ref{prop1} and \ref{prop_opt_sound}.
\end{proof}

%% file: D-domains.tex
\section{Experiments}
\label{sec:exp_appendix}

\subsection{Experimental domains}
We describe the characteristics of the domains and tasks used in our experiments in Tables \ref{tab:dataset} and \ref{tab:tasks}. Figure \ref{fig:sols} shows example solutions for some of the tasks.

\begin{table}
\footnotesize
\centering
\begin{tabular}{@{}l|cccc@{}}
\textbf{Task} & \textbf{\# examples} & \textbf{\# relations} & \textbf{\# constants} & \textbf{\# facts}\\
\midrule
\emph{trains} & 1000 & 20 & 8561 & 28503 \\
\midrule
\emph{zendo1} & 100 & 16 & 1049 & 2270\\
\emph{zendo2} & 100 & 16 & 1047 & 2184\\
\emph{zendo3} & 100 & 16 & 1100 & 2320\\
\emph{zendo4} & 100 & 16 & 987 & 2087\\
\midrule
\emph{imdb1} & 383 & 6 & 299 & 1330 \\
\emph{imdb2} & 71825 & 6 & 299 & 1330 \\
\emph{imdb3} & 121801 & 6 & 299 & 1330 \\
\midrule
\emph{krk1} & 20 & 10 & 108 & 4223 \\
\emph{krk2} & 20 & 10 & 108 & 168 \\
\emph{krk3} & 20 & 8 & 108 & 140 \\
\midrule
\emph{md} & 54 & 12 & 13 & 29 \\
\emph{buttons} & 530 & 13 & 60 & 656 \\
\emph{buttons-g} & 44 & 21 & 31 & 130 \\
\emph{rps} & 464 & 6 & 64 & 405 \\
\emph{coins} & 2544 & 9 & 110 & 1101 \\
\emph{coins-g} & 125 & 118 & 68 & 742 \\
\emph{attrition} & 672 & 12 & 65 & 163\\
\emph{centipede} & 26 & 34 & 61 & 138 
\\\midrule
\emph{adj\_red} & 40 & 5 & 1034 & 18256\\
\emph{connected} & 40 & 5 & 1030 & 4602\\
\emph{cyclic} & 40 & 5 & 1012 & 4426\\
\emph{colouring} & 40 & 5 & 662 & 10021\\
\emph{undirected} & 40 & 5 & 998 & 17272\\
\emph{2children} & 40 & 6 & 672 & 11172
\\\midrule
\emph{contains} & 20 & 10 & $\infty$ & $\infty$\\
\emph{dropk} & 20 & 10 & $\infty$ & $\infty$\\
\emph{droplast} & 20 & 10 & $\infty$ & $\infty$\\
\emph{evens} & 20 & 10 & $\infty$ & $\infty$\\
\emph{finddup} & 20 & 10 & $\infty$ & $\infty$\\
\emph{last} & 20 & 10 & $\infty$ & $\infty$\\
\emph{len} & 20 & 10 & $\infty$ & $\infty$\\
\emph{reverse} & 20 & 10 & $\infty$ & $\infty$\\
\emph{sorted} & 20 & 10 & $\infty$ & $\infty$\\
\emph{sumlist} & 20 & 10 & $\infty$ & $\infty$\\
\end{tabular}
\caption{
Experimental domain description.
}

\label{tab:dataset}
\end{table}

\begin{table}
\footnotesize
\centering
\begin{tabular}{@{}l|ccp{5mm}c@{}}
\textbf{Task} & \textbf{\#rules} & \textbf{\#literals} & \textbf{max rule size} & \textbf{recursion}\\
\\
\midrule
\emph{train1} & 1 & 6 &  6 & no\\
\emph{train2} & 2 & 11 & 6 & no\\
\emph{train3} & 3 & 17 &  7& no\\
\emph{train4} & 4 & 26 &  7& no\\
\midrule
\emph{zendo1} & 1 & 7 & 7 & no\\
\emph{zendo2} & 2 & 14 & 7 & no\\
\emph{zendo3} & 3 & 20 & 7& no\\
\emph{zendo4} & 4 & 23 & 7& no\\
\midrule
\emph{imdb1} & 1 & 5 & 5 & no\\
\emph{imdb2} & 1 & 5 & 5 & no\\
\emph{imdb3} & 2 & 10 & 5 & no\\
\midrule
\emph{krk1} & 1 & 8 & 8 & no\\
\emph{krk2} & 1 & 36 & 9 & no\\
\emph{krk3} & 2 & 16 & 8 & no\\
\midrule
\emph{md} & 2 & 11 & 6 & no\\
\emph{buttons} & 10 & 61 & 7& no\\
\emph{rps} & 4 & 25 & 7& no\\
\emph{coins} & 16 & 45 & 7& no\\
\emph{attrition} & 3 & 14 & 5& no\\
\emph{centipede} & 2 & 8 & 4& no\\
\midrule
\emph{adj\_red} & 1 & 4 & 4& no\\
\emph{connected} & 2 & 5 & 3& yes\\
\emph{cyclic} & 3 & 7 & 3& yes\\
\emph{colouring} & 1 & 4 & 4& no\\
\emph{undirected} & 2 & 4 & 2& no\\
\emph{2children} & 1 & 4 & 4& no\\
\midrule
\emph{contains} & 3 & 9 & 3& yes\\
\emph{dropk} & 2 & 7 & 4& yes\\
\emph{droplast} & 2 & 8 & 5& yes\\
\emph{evens} & 2 & 7 & 5& yes\\
\emph{finddup} & 2 & 7 & 4& yes\\
\emph{last} & 2 & 7 & 4& yes\\
\emph{len} & 2 & 7 & 4& yes\\
\emph{reverse} & 2 & 8 & 5& yes\\
\emph{sorted} & 2 & 9 & 6& yes\\
\emph{sumlist} & 2 & 7 & 5& yes\\
\end{tabular}
\caption{
Statistics about the optimal solutions for each task.
For instance, the optimal solution for \emph{buttons} has 10 rules and 61 literals and the largest rule has 7 literals.
}

\label{tab:tasks}
\end{table}

\paragraph{Michalski trains.}
The goal of these tasks is to find a hypothesis that distinguishes eastbound and westbound trains \cite{michalski:trains}. There are four increasingly complex tasks. 
There are 1000 examples but the distribution of positive and negative examples is different for each task.
We randomly sample the examples and split them into 80/20 train/test partitions.

\paragraph{Zendo.} Zendo is an inductive game in which one player, the Master, creates a rule for structures made of pieces with varying attributes to follow. The other players, the Students, try to discover the rule by building and studying structures which are labelled by the Master as following or breaking the rule. The first student to correctly state the rule wins. We learn four increasingly complex rules for structures made of at most 5 pieces of varying color, size, orientation and position. 
Zendo is a challenging game that has attracted much interest in cognitive science \cite{zendo}.

\paragraph{IMDB.}
The real-world IMDB dataset \cite{mihalkova2007} includes relations between movies, actors, directors, movie genre, and gender. It has been created from the International Movie Database (IMDB.com) database. We learn the relation \emph{workedunder/2}, a more complex variant \emph{workedwithsamegender/2}, and the disjunction of the two.
This dataset is frequently used \cite{alps}.

\paragraph{Chess.} The task is to learn chess patterns in the king-rook-king (\emph{krk}) endgame, which is the chess ending with white having a king and a rook and black having a king. We learn the concept of white rook protection by the white king (\emph{krk1}) \cite{celine:bottom}, king opposition (\emph{krk2}) and rook attack (\emph{krk3}).

\paragraph{IGGP.}
In \emph{inductive general game playing} (IGGP)  \cite{iggp} the task is to induce a hypothesis to explain game traces from the general game playing competition \cite{ggp}.
IGGP is notoriously difficult for machine learning approaches.
The currently best-performing system can only learn perfect solutions for 40\% of the tasks.
Moreover, although seemingly a toy problem, IGGP is representative of many real-world problems, such as inducing semantics of programming languages \cite{DBLP:conf/ilp/BarthaC19}. 
We use six games: \emph{minimal decay (md)}, \emph{buttons}, \emph{rock - paper - scissors (rps)}, \emph{coins}, \emph{attrition}, and \emph{centipede}.


\paragraph{Graph problems.} This dataset involves solving usual graph problems \cite{dilp,DBLP:conf/icml/GlanoisJFWZ0LH22}. Graphs are represented by sets of edges linking coloured nodes.

\paragraph{Program Synthesis.} This dataset includes list transformation tasks. It involves learning recursive programs which have been identified as a difficult challenge for ILP systems \cite{ilp20}. 

\subsection{Experimental Setup}
We measure the mean and standard error of the predictive accuracy and learning time.
We use a 3.8 GHz 8-Core Intel Core i7 with 32GB of ram.
The systems use a single CPU.



\subsection{Experimental Results}

\subsubsection{Comparison against other ILP systems}

To answer \textbf{Q2} we compare \name{} against \popper{}, \dcc{}, \ale{}, and \metagol{}, which we describe below.

\begin{description}
\item[\popper{} and \dcc{}] \popper{} and \dcc{} use identical biases to \name{} so the comparison is direct, i.e. fair.
We use \popper{} 1.1.0\footnote{\url{https://github.com/logic-and-learning-lab/Popper/releases/tag/v1.1.0}}.

\item[\ale{}] \ale{} excels at learning many large non-recursive rules and should excel at the trains and IGGP tasks.
Although \ale{} can learn recursive programs, it struggles to do so.
\name{} and \ale{} use similar biases so the comparison can be considered reasonably fair.
For instance, Figures \ref{fig:biasname} and \ref{fig:biasale} show the \name{} and \ale{} biases used for the Zendo tasks.

\item[\metagol{}] \metagol{} is one of the few systems that can learn recursive Prolog programs.
\metagol{} uses user-provided \emph{metarules} (program templates) to guide the search for a solution.
We use an approximate universal set of metarules \cite{reduce}.
However, these metarules are only sufficient to learn programs with literals of arity at most two, and so \metagol{} cannot solve many tasks we consider.



\end{description}
\noindent
We also tried/considered other ILP systems.
We considered \ilasp{} \cite{ilasp}.
However, \ilasp{} builds on \aspal{} and first precomputes every possible rule in a hypothesis space, which is infeasible for our datasets.
In addition, \ilasp{} cannot learn Prolog programs so is unusable in the synthesis tasks.
For instance, it would require precomputing $10^{15}$ rules for the \emph{coins} task.
We also considered \textsc{hexmil} \cite{hexmil} and \textsc{louise} \cite{louise}, both metarule-based approaches similar to \metagol{}.
However, their performance was considerably worse than \metagol{} so we have excluded them from the comparison.

\begin{figure}
\begin{lstlisting}[caption=\name{}]
max_clause(6).
max_vars(6).
max_body(6).

head_pred(zendo,1).
body_pred(piece,2).
body_pred(contact,2).
body_pred(coord1,2).
body_pred(coord2,2).
body_pred(size,2).
body_pred(blue,1).
body_pred(green,1).
body_pred(red,1).
body_pred(small,1).
body_pred(medium,1).
body_pred(large,1).
body_pred(upright,1).
body_pred(lhs,1).
body_pred(rhs,1).
body_pred(strange,1).

type(zendo,(state,)).
type(piece,(state,piece)).
type(contact,(piece,piece)).
type(coord1,(piece,real)).
type(coord2,(piece,real)).
type(size,(piece,real)).
type(blue,(piece,)).
type(green,(piece,)).
type(red,(piece,)).
type(small,(real,)).
type(medium,(real,)).
type(large,(real,)).
type(upright,(piece,)).
type(lhs,(piece,)).
type(rhs,(piece,)).
type(strange,(piece,)).

direction(zendo,(in,)).
direction(piece,(in,out)).
direction(contact,(in,out)).
direction(coord1,(in,out)).
direction(coord2,(in,out)).
direction(size,(in,out)).
direction(blue,(in,)).
direction(green,(in,)).
direction(red,(in,)).
direction(small,(in,)).
direction(medium,(in,)).
direction(large,(in,)).
direction(upright,(in,)).
direction(lhs,(in,)).
direction(rhs,(in,)).
direction(strange,(in,)).
\end{lstlisting}
\caption{The bias files for \name{} for the learning tasks \emph{zendo}.}
\label{fig:biasname}
\end{figure}

\begin{figure}
\begin{lstlisting}[caption=\ale{}]
:- aleph_set(i,6).
:- aleph_set(clauselength,7).
:- aleph_set(nodes,50000).

:- modeh(*,zendo(+state)).
:- modeb(*,piece(+state,-piece)).
:- modeb(*,contact(+piece,-piece)).
:- modeb(*,coord1(+piece,-real)).
:- modeb(*,coord2(+piece,-real)).
:- modeb(*,size(+piece,-real)).
:- modeb(*,blue(+piece)).
:- modeb(*,green(+piece)).
:- modeb(*,red(+piece)).
:- modeb(*,small(+real)).
:- modeb(*,medium(+real)).
:- modeb(*,large(+real)).
:- modeb(*,upright(+piece)).
:- modeb(*,lhs(+piece)).
:- modeb(*,rhs(+piece)).
:- modeb(*,strange(+piece)).

\end{lstlisting}
\caption{The bias files for \ale{} for the learning tasks \emph{zendo}.}
\label{fig:biasale}
\end{figure}

\paragraph{Results.}
Table \ref{tab:q1-600} shows the predictive accuracies of the systems when given a \textbf{60 minutes} timeout.
Table \ref{tab:q1times} shows the termination times of the systems.
Figure \ref{fig:sols} shows example solutions.

\begin{table*}
\centering
\small
\begin{tabular}{@{}l|ccccc@{}}
\textbf{Task} & \textbf{\name{}} & \textbf{\textsc{\popper{}}} & \textbf{\dcc{}} & \textbf{\ale{}} & \textbf{\metagol{}}\\
\midrule
\emph{trains1} & \textbf{100 $\pm$ 0}  & \textbf{100 $\pm$ 0}  & \textbf{100 $\pm$ 0}  & \textbf{100 $\pm$ 0}  & 27 $\pm$ 0 \\
\emph{trains2} & 98 $\pm$ 0 & 98 $\pm$ 0 & 98 $\pm$ 0 & \textbf{100 $\pm$ 0}  & 49 $\pm$ 12 \\
\emph{trains3} & \textbf{100 $\pm$ 0}  & 79 $\pm$ 0 & \textbf{100 $\pm$ 0}  & \textbf{100 $\pm$ 0}  & 79 $\pm$ 0 \\
\emph{trains4} & \textbf{100 $\pm$ 0}  & 32 $\pm$ 0 & \textbf{100 $\pm$ 0}  & \textbf{100 $\pm$ 0}  & 32 $\pm$ 0 \\
\midrule
\emph{zendo1} & \textbf{97 $\pm$ 0}  & \textbf{97 $\pm$ 0}  & \textbf{97 $\pm$ 0}  & 90 $\pm$ 2 & 73 $\pm$ 7 \\
\emph{zendo2} & \textbf{93 $\pm$ 2}  & 50 $\pm$ 0 & 81 $\pm$ 3 & \textbf{93 $\pm$ 3}  & 50 $\pm$ 0 \\
\emph{zendo3} & \textbf{95 $\pm$ 2}  & 50 $\pm$ 0 & 78 $\pm$ 3 & \textbf{95 $\pm$ 2}  & 50 $\pm$ 0 \\
\emph{zendo4} & \textbf{93 $\pm$ 1}  & 54 $\pm$ 4 & 88 $\pm$ 1 & 88 $\pm$ 1 & 57 $\pm$ 5 \\
\midrule
\emph{imdb1} & \textbf{100 $\pm$ 0}  & \textbf{100 $\pm$ 0}  & \textbf{100 $\pm$ 0}  & \textbf{100 $\pm$ 0}  & 16 $\pm$ 0 \\
\emph{imdb2} & \textbf{100 $\pm$ 0}  & \textbf{100 $\pm$ 0}  & \textbf{100 $\pm$ 0}  & 50 $\pm$ 0 & 50 $\pm$ 0 \\
\emph{imdb3} & \textbf{100 $\pm$ 0}  & 50 $\pm$ 0 & \textbf{100 $\pm$ 0}  & 50 $\pm$ 0 & 50 $\pm$ 0 \\
\midrule
\emph{krk1} & \textbf{98 $\pm$ 0}  & \textbf{98 $\pm$ 0}  & \textbf{98 $\pm$ 0}  & 97 $\pm$ 0 & 50 $\pm$ 0 \\
\emph{krk2} & 79 $\pm$ 4 & 50 $\pm$ 0 & 54 $\pm$ 4 & \textbf{95 $\pm$ 0}  & 50 $\pm$ 0 \\
\emph{krk3} & 54 $\pm$ 0 & 50 $\pm$ 0 & 50 $\pm$ 0 & \textbf{90 $\pm$ 4}  & 50 $\pm$ 0 \\
\midrule
\emph{md} & \textbf{100 $\pm$ 0}  & 37 $\pm$ 13 & \textbf{100 $\pm$ 0}  & 94 $\pm$ 0 & 11 $\pm$ 0 \\
\emph{buttons} & \textbf{100 $\pm$ 0}  & 19 $\pm$ 0 & \textbf{100 $\pm$ 0}  & 96 $\pm$ 0 & 19 $\pm$ 0 \\
\emph{rps} & \textbf{100 $\pm$ 0}  & 18 $\pm$ 0 & \textbf{100 $\pm$ 0}  & \textbf{100 $\pm$ 0}  & 18 $\pm$ 0 \\
\emph{coins} & \textbf{100 $\pm$ 0}  & 17 $\pm$ 0 & \textbf{100 $\pm$ 0}  & 17 $\pm$ 0 & 17 $\pm$ 0 \\
\emph{buttons-g} & \textbf{100 $\pm$ 0}  & 50 $\pm$ 0 & 86 $\pm$ 1 & \textbf{100 $\pm$ 0}  & 50 $\pm$ 0 \\
\emph{coins-g} & \textbf{100 $\pm$ 0}  & 50 $\pm$ 0 & 90 $\pm$ 6 & \textbf{100 $\pm$ 0}  & 50 $\pm$ 0 \\
\emph{attrition} & \textbf{98 $\pm$ 0}  & 2 $\pm$ 0 & 2 $\pm$ 0 & \textbf{98 $\pm$ 0}  & 2 $\pm$ 0 \\
\emph{centipede} & \textbf{100 $\pm$ 0}  & \textbf{100 $\pm$ 0}  & 81 $\pm$ 6 & \textbf{100 $\pm$ 0}  & 50 $\pm$ 0 \\
\midrule
\emph{adj\_red} & \textbf{100 $\pm$ 0}  & \textbf{100 $\pm$ 0}  & \textbf{100 $\pm$ 0}  & 50 $\pm$ 0 & \textbf{100 $\pm$ 0}  \\
\emph{connected} & \textbf{98 $\pm$ 0}  & 81 $\pm$ 7 & 82 $\pm$ 7 & 51 $\pm$ 0 & 50 $\pm$ 0 \\
\emph{cyclic} & \textbf{89 $\pm$ 3}  & 80 $\pm$ 7 & 85 $\pm$ 5 & 50 $\pm$ 0 & 50 $\pm$ 0 \\
\emph{colouring} & \textbf{98 $\pm$ 1}  & \textbf{98 $\pm$ 1}  & \textbf{98 $\pm$ 1}  & 50 $\pm$ 0 & \textbf{98 $\pm$ 1}  \\
\emph{undirected} & \textbf{100 $\pm$ 0}  & \textbf{100 $\pm$ 0}  & \textbf{100 $\pm$ 0}  & 50 $\pm$ 0 & 50 $\pm$ 0 \\
\emph{2children} & \textbf{100 $\pm$ 0}  & 99 $\pm$ 0 & \textbf{100 $\pm$ 0}  & 50 $\pm$ 0 & 94 $\pm$ 4 \\
\midrule
\emph{dropk} & \textbf{100 $\pm$ 0}  & \textbf{100 $\pm$ 0}  & \textbf{100 $\pm$ 0}  & 55 $\pm$ 4 & 50 $\pm$ 0 \\
\emph{droplast} & \textbf{100 $\pm$ 0}  & 95 $\pm$ 5 & \textbf{100 $\pm$ 0}  & 50 $\pm$ 0 & 50 $\pm$ 0 \\
\emph{evens} & \textbf{100 $\pm$ 0}  & \textbf{100 $\pm$ 0}  & \textbf{100 $\pm$ 0}  & 50 $\pm$ 0 & 50 $\pm$ 0 \\
\emph{finddup} & \textbf{99 $\pm$ 0}  & 98 $\pm$ 0 & \textbf{99 $\pm$ 0}  & 50 $\pm$ 0 & 50 $\pm$ 0 \\
\emph{last} & \textbf{100 $\pm$ 0}  & \textbf{100 $\pm$ 0}  & \textbf{100 $\pm$ 0}  & 55 $\pm$ 4 & \textbf{100 $\pm$ 0}  \\
\emph{contains} & \textbf{100 $\pm$ 0}  & \textbf{100 $\pm$ 0}  & 99 $\pm$ 0 & 56 $\pm$ 2 & 50 $\pm$ 0 \\
\emph{len} & \textbf{100 $\pm$ 0}  & \textbf{100 $\pm$ 0}  & \textbf{100 $\pm$ 0}  & 50 $\pm$ 0 & 50 $\pm$ 0 \\
\emph{reverse} & \textbf{100 $\pm$ 0}  & 85 $\pm$ 7 & \textbf{100 $\pm$ 0}  & 50 $\pm$ 0 & 50 $\pm$ 0 \\
\emph{sorted} & \textbf{100 $\pm$ 0}  & \textbf{100 $\pm$ 0}  & \textbf{100 $\pm$ 0}  & 74 $\pm$ 2 & 50 $\pm$ 0 \\
\emph{sumlist} & \textbf{100 $\pm$ 0}  & \textbf{100 $\pm$ 0}  & \textbf{100 $\pm$ 0}  & 50 $\pm$ 0 & \textbf{100 $\pm$ 0}  \\
\end{tabular}
\caption{
Predictive accuracies with a 60-minute learning timeout.
We round accuracies to integer values. The error is the standard deviation.
}
\label{tab:q1-600}
\end{table*}

\begin{table*}
\small
\centering
\begin{tabular}{@{}l|ccccc@{}}
\textbf{Task} & \textbf{\name{}} & \textbf{\textsc{\popper{}}} & \textbf{\dcc{}} & \textbf{\ale{}} & \textbf{\metagol{}}\\
\midrule
\emph{trains1} & 4 $\pm$ 0.1 & 5 $\pm$ 0.4 & 8 $\pm$ 0.7 & \textbf{3 $\pm$ 0.6}  & \emph{timeout} \\
\emph{trains2} & 4 $\pm$ 0.1 & 82 $\pm$ 25 & 10 $\pm$ 0.9 & \textbf{2 $\pm$ 0.2}  & 3010 $\pm$ 261 \\
\emph{trains3} & 18 $\pm$ 0.5 & \emph{timeout} & \emph{timeout} & \textbf{13 $\pm$ 3}  & \emph{timeout} \\
\emph{trains4} & \textbf{16 $\pm$ 0.5}  & \emph{timeout} & \emph{timeout} & 136 $\pm$ 55 & \emph{timeout} \\
\midrule
\emph{zendo1} & 3 $\pm$ 0.6 & 7 $\pm$ 1 & 7 $\pm$ 1 & \textbf{1 $\pm$ 0.3}  & 1801 $\pm$ 599 \\
\emph{zendo2} & 49 $\pm$ 5 & \emph{timeout} & 3256 $\pm$ 345 & \textbf{1 $\pm$ 0.2}  & \emph{timeout} \\
\emph{zendo3} & 55 $\pm$ 6 & \emph{timeout} & \emph{timeout} & \textbf{1 $\pm$ 0.1}  & \emph{timeout} \\
\emph{zendo4} & 53 $\pm$ 11 & 3243 $\pm$ 359 & 2939 $\pm$ 444 & \textbf{1 $\pm$ 0.3}  & 2883 $\pm$ 478 \\
\midrule
\emph{imdb1} & \textbf{2 $\pm$ 0.1}  & 3 $\pm$ 0.2 & 3 $\pm$ 0.2 & 142 $\pm$ 41 & \emph{timeout} \\
\emph{imdb2} & \textbf{3 $\pm$ 0.1}  & 11 $\pm$ 1 & \textbf{3 $\pm$ 0.2}  & \emph{timeout} & \emph{timeout} \\
\emph{imdb3} & \textbf{547 $\pm$ 46}  & 875 $\pm$ 166 & 910 $\pm$ 320 & \emph{timeout} & \emph{timeout} \\
\midrule
\emph{krk1} & 28 $\pm$ 6 & 1358 $\pm$ 321 & 188 $\pm$ 53 & \textbf{3 $\pm$ 1}  & 79 $\pm$ 6 \\
\emph{krk2} & 3459 $\pm$ 141 & \emph{timeout} & \emph{timeout} & \textbf{11 $\pm$ 4}  & 97 $\pm$ 6 \\
\emph{krk3} & \emph{timeout} & \emph{timeout} & \emph{timeout} & \textbf{16 $\pm$ 3}  & 88 $\pm$ 6 \\
\midrule
\emph{md} & 13 $\pm$ 1 & 3357 $\pm$ 196 & \emph{timeout} & \textbf{4 $\pm$ 0}  & \emph{timeout} \\
\emph{buttons} & \textbf{23 $\pm$ 3}  & \emph{timeout} & \emph{timeout} & 99 $\pm$ 0.1 & \emph{timeout} \\
\emph{rps} & 87 $\pm$ 15 & \emph{timeout} & \emph{timeout} & 20 $\pm$ 0 & \textbf{0.1 $\pm$ 0}  \\
\emph{coins} & 490 $\pm$ 35 & \emph{timeout} & \emph{timeout} & \emph{timeout} & \textbf{0.2 $\pm$ 0}  \\
\emph{buttons-g} & 3 $\pm$ 0.1 & \emph{timeout} & \emph{timeout} & 86 $\pm$ 0.2 & \textbf{0.1 $\pm$ 0}  \\
\emph{coins-g} & 105 $\pm$ 6 & \emph{timeout} & \emph{timeout} & 9 $\pm$ 0.1 & \textbf{0.1 $\pm$ 0}  \\
\emph{attrition} & 26 $\pm$ 1 & \emph{timeout} & \emph{timeout} & 678 $\pm$ 25 & \textbf{2 $\pm$ 0}  \\
\emph{centipede} & 9 $\pm$ 0.4 & 1102 $\pm$ 136 & 2104 $\pm$ 501 & 12 $\pm$ 0 & \textbf{2 $\pm$ 0}  \\
\midrule
\emph{adj\_red} & 2 $\pm$ 0.1 & 5 $\pm$ 0 & 6 $\pm$ 0.2 & 479 $\pm$ 349 & \textbf{0.5 $\pm$ 0}  \\
\emph{connected} & \textbf{5 $\pm$ 1}  & 112 $\pm$ 71 & 735 $\pm$ 478 & 435 $\pm$ 353 & \emph{timeout} \\
\emph{cyclic} & \textbf{35 $\pm$ 13}  & 1321 $\pm$ 525 & 1192 $\pm$ 456 & 1120 $\pm$ 541 & \emph{timeout} \\
\emph{colouring} & 2 $\pm$ 0.1 & 6 $\pm$ 0.1 & 5 $\pm$ 0.2 & 2373 $\pm$ 518 & \textbf{0.4 $\pm$ 0.1}  \\
\emph{undirected} & \textbf{2 $\pm$ 0.1}  & 6 $\pm$ 0.1 & 6 $\pm$ 0.2 & 227 $\pm$ 109 & \emph{timeout} \\
\emph{2children} & \textbf{2 $\pm$ 0.1}  & 7 $\pm$ 0.2 & 6 $\pm$ 0.3 & 986 $\pm$ 405 & 360 $\pm$ 360 \\
\midrule
\emph{dropk} & 7 $\pm$ 3 & 17 $\pm$ 2 & 14 $\pm$ 2 & 4 $\pm$ 1 & \textbf{0.1 $\pm$ 0}  \\
\emph{droplast} & \textbf{3 $\pm$ 0.1}  & 372 $\pm$ 359 & 13 $\pm$ 1 & 763 $\pm$ 67 & 3260 $\pm$ 340 \\
\emph{evens} & 3 $\pm$ 0.1 & 29 $\pm$ 3 & 25 $\pm$ 2 & \textbf{2 $\pm$ 0.2}  & 2713 $\pm$ 466 \\
\emph{finddup} & 11 $\pm$ 5 & 136 $\pm$ 14 & 149 $\pm$ 7 & \textbf{0.8 $\pm$ 0.1}  & 2894 $\pm$ 471 \\
\emph{last} & 2 $\pm$ 0.2 & 12 $\pm$ 0.8 & 11 $\pm$ 0.6 & 2 $\pm$ 0.3 & \textbf{1 $\pm$ 0.3}  \\
\emph{contains} & \textbf{17 $\pm$ 0.8}  & 299 $\pm$ 52 & 158 $\pm$ 48 & 64 $\pm$ 5 & \emph{timeout} \\
\emph{len} & 3 $\pm$ 0.2 & 52 $\pm$ 5 & 45 $\pm$ 2 & \textbf{2 $\pm$ 0.2}  & \emph{timeout} \\
\emph{reverse} & 40 $\pm$ 5 & 1961 $\pm$ 401 & 1924 $\pm$ 300 & \textbf{3 $\pm$ 0.1}  & 2331 $\pm$ 524 \\
\emph{sorted} & 127 $\pm$ 78 & 111 $\pm$ 11 & 131 $\pm$ 10 & \textbf{1 $\pm$ 0.1}  & 2687 $\pm$ 471 \\
\emph{sumlist} & 4 $\pm$ 0.1 & 256 $\pm$ 27 & 221 $\pm$ 12 & \textbf{0.3 $\pm$ 0}  & \textbf{0.3 $\pm$ 0}  \\

\end{tabular}
\caption{
Learning times with a 60-minute timeout.
A \emph{timeout} entry means that the system did not terminate in the given time.
We round times over one second to the nearest second.
The error is standard deviation.
}
\label{tab:q1times}
\end{table*}

\begin{figure*}
\centering
\footnotesize
\begin{lstlisting}[caption=trains2\label{trains2}]
east(A):-car(A,C),roof_open(C),load(C,B),triangle(B).
east(A):-car(A,C),car(A,B),roof_closed(B),two_wheels(C),roof_open(C).
\end{lstlisting}

\centering
\footnotesize
\begin{lstlisting}[caption=trains4\label{trains4}]
east(A):-has_car(A,D),has_load(D,B),has_load(D,C),rectangle(B),diamond(C).
east(A):-has_car(A,B),has_load(B,C),hexagon(C),roof_open(B),three_load(C).
east(A):-has_car(A,E),has_car(A,D),has_load(D,C),triangle(C),has_load(E,B),hexagon(B).
east(A):-has_car(A,C),roof_open(C),has_car(A,B),roof_flat(B),short(C),long(B).
\end{lstlisting}
\centering

\begin{lstlisting}[caption=zendo1\label{zendo1}]
zendo1(A):- piece(A,C),size(C,B),blue(C),small(B),contact(C,D),red(D).
\end{lstlisting}
\centering

\begin{lstlisting}[caption=zendo2\label{zendo2}]
zendo2(A):- piece(A,B),piece(A,D),piece(A,C),green(D),red(B),blue(C).
zendo2(A):- piece(A,D),piece(A,B),coord1(B,C),green(D),lhs(B),coord1(D,C).
\end{lstlisting}
\centering

\begin{lstlisting}[caption=zendo3\label{zendo3}]
zendo3(A):- piece(A,D),blue(D),coord1(D,B),piece(A,C),coord1(C,B),red(C).
zendo3(A):- piece(A,D),contact(D,C),rhs(D),size(C,B),large(B).
zendo3(A):- piece(A,B),upright(B),contact(B,D),blue(D),size(D,C),large(C).
\end{lstlisting}
\centering

\begin{lstlisting}[caption=zendo4\label{zendo4}]
zendo4(A):- piece(A,C),contact(C,B),strange(B),upright(C).
zendo4(A):- piece(A,D),contact(D,C),coord2(C,B),coord2(D,B).
zendo4(A):- piece(A,D),contact(D,C),size(C,B),red(D),medium(B).
zendo4(A):- piece(A,D),blue(D),lhs(D),piece(A,C),size(C,B),small(B).\end{lstlisting}
\centering

\begin{lstlisting}[caption=krk1\label{krk1}]
f(A):-cell(A,B,C,D),white(C),cell(A,E,C,F),rook(D),king(F),distance(B,E,G),one(G).
\end{lstlisting}

\begin{lstlisting}[caption=krk2\label{krk2}]
f(A):-cell(A,B,C,D),white(C),cell(A,E,F,D),king(D),black(F),samerank(B,E),nextfile(B,G),nextfile(G,E).
f(A):-cell(A,B,C,D),white(C),cell(A,E,F,D),king(D),black(F),samerank(B,E),nextfile(E,G),nextfile(G,B).
f(A):-cell(A,B,C,D),white(C),cell(A,E,F,D),king(D),black(F),samefile(B,E),nextrank(B,G),nextrank(G,E).
f(A):-cell(A,B,C,D),white(C),cell(A,E,F,D),king(D),black(F),samefile(B,E),nextrank(F,G),nextrank(G,B).
\end{lstlisting}

\begin{lstlisting}[caption=krk3\label{krk3}]
f(A):-cell(A,B,C,D),white(C),cell(A,E,F,G),rook(D),king(G),black(F),samerank(B,E).
f(A):-cell(A,B,C,D),white(C),cell(A,E,F,G),rook(D),king(G),black(F),samefile(B,E).
\end{lstlisting}

\begin{lstlisting}[caption=minimal decay\label{minimaldecay}]
next_value(A,B):-c_player(D),c_pressButton(C),c5(B),does(A,D,C).
next_value(A,B):-c_player(C),my_true_value(A,E),does(A,C,D),my_succ(B,E),c_noop(D).
\end{lstlisting}

\begin{lstlisting}[caption=buttons \label{buttons}]
next(A,B):-c_p(B),c_c(C),does(A,D,C),my_true(A,B),my_input(D,C).
next(A,B):-my_input(C,E),c_p(D),my_true(A,D),c_b(E),does(A,C,E),c_q(B).
next(A,B):-my_input(C,D),not_my_true(A,B),does(A,C,D),c_p(B),c_a(D).
next(A,B):-c_a(C),does(A,D,C),my_true(A,B),c_q(B),my_input(D,C).
next(A,B):-my_input(C,E),c_p(B),my_true(A,D),c_b(E),does(A,C,E),c_q(D).
next(A,B):-c_c(D),my_true(A,C),c_r(B),role(E),does(A,E,D),c_q(C).
next(A,B):-my_true(A,C),my_succ(C,B).
next(A,B):-my_input(C,D),does(A,C,D),my_true(A,B),c_r(B),c_b(D).
next(A,B):-my_input(C,D),does(A,C,D),my_true(A,B),c_r(B),c_a(D).
next(A,B):-my_true(A,E),c_c(C),does(A,D,C),c_q(B),c_r(E),my_input(D,C).
\end{lstlisting}
\begin{lstlisting}[caption=rps\label{rps}]
next_score(A,B,C):-does(A,B,E),different(G,B),my_true_score(A,B,F),beats(E,D),my_succ(F,C),
                   does(A,G,D).
next_score(A,B,C):-different(G,B),beats(D,F),my_true_score(A,E,C),does(A,G,D),does(A,E,F).
next_score(A,B,C):-my_true_score(A,B,C),does(A,B,D),does(A,E,D),different(E,B).
\end{lstlisting}

\caption{Example solutions.}
\end{figure*}

\setcounter{figure}{2}
\begin{figure*}
\footnotesize
\begin{lstlisting}[caption=coins\label{coins}]
next_cell(A,B,C):-does_jump(A,E,F,D),role(E),different(B,D),my_true_cell(A,B,C),different(F,B).
next_cell(A,B,C):-my_pos(E),role(D),c_zerocoins(C),does_jump(A,D,B,E).
next_cell(A,B,C):-role(D),does_jump(A,D,E,B),c_twocoins(C),different(B,E).
next_cell(A,B,C):-does_jump(A,F,E,D),role(F),my_succ(E,B),my_true_cell(A,B,C),different(E,D).
\end{lstlisting}

\begin{lstlisting}[caption=coins-goal\label{coins-goal}]
goal(A,B,C):- role(B),pos_5(D),my_true_step(A,D),score_100(C).
goal(A,B,C):- c_onecoin(E),my_pos(D),my_true_cell(A,D,E),score_0(C),role(B).
\end{lstlisting}

\begin{lstlisting}[caption=adjacent$\_$to$\_$red\label{adjacent_to_red}]
f(A):- edge(A,C),colour(C,B),red(B).
\end{lstlisting}

\begin{lstlisting}[caption=connected\label{connectedness}]
f(A,B):- edge(A,B).
f(A,B):- edge(A,C),f(C,B).
\end{lstlisting}

\begin{lstlisting}[caption=cyclic\label{cyclic}]
f(A):- f_1(A,A).
f_1(A,B):- edge(A,B).
f_1(A,B):- edge(A,C),f_1(C,B).
\end{lstlisting}

\begin{lstlisting}[caption=graph$\_$colouring\label{graph_colouring}]
f(A):- edge(A,C),colour(C,B),colour(A,B).
\end{lstlisting}

\begin{lstlisting}[caption=undirected$\_$edge\label{undirected_edge}]
f(A,B):- edge(A,B).
f(A,B):- edge(B,A).
\end{lstlisting}

\begin{lstlisting}[caption=two$\_$children\label{two_children}]
f(A):- edge(A,B),edge(A,C),diff(B,C).
\end{lstlisting}

\centering
\begin{lstlisting}[caption=sorted]
sorted(A):-tail(A,B),empty(B).
sorted(A):-tail(A,D),head(A,B),head(D,C),geq(C,B),sorted(D).
\end{lstlisting}

\centering
\begin{lstlisting}[caption=contains]
contains(A):- head(A,B),c_6(B).
contains(A):- head(A,B),c_9(B).
contains(A):- tail(A,B),contains(B).
\end{lstlisting}

\centering
\begin{lstlisting}[caption=reverse]
reverse(A,B):- empty_out(B),empty(A).
reverse(A,B):- head(A,D),tail(A,E),reverse(E,C),append(C,D,B).
\end{lstlisting}

\centering
\begin{lstlisting}[caption=sumlist]
sumlist(A,B):- head(A,B).
sumlist(A,B):- head(A,D),tail(A,C),sumlist(C,E),sum(D,E,B).
\end{lstlisting}

\centering
\begin{lstlisting}[caption=dropk]
dropk(A,B,C):- tail(A,C),odd(B),one(B).
dropk(A,B,C):- decrement(B,E),tail(A,D),dropk(D,E,C).
\end{lstlisting}


\caption{Example solutions.}
\label{fig:sols}
\end{figure*}

\section{Example \name{} Output}
\label{sec:outputs}
Figure \ref{fig:coinsg} shows an example \name{} output on the \emph{coins-goal} task.
Figure \ref{fig:buttons-goa} shows an example \name{} output on the \emph{buttons-goal} task.

\begin{figure*}
\begin{lstlisting}
Num. pos examples: 62
Num. neg examples: 62
Searching programs of size: 4
Searching programs of size: 5
********************
New best hypothesis:
tp:1 fn:61 size:5
goal(A,B,C):- score_0(C),role(B),pos_1(D),my_true_step(A,D).
********************
********************
New best hypothesis:
tp:3 fn:59 size:10
goal(A,B,C):- score_0(C),role(B),pos_1(D),my_true_step(A,D).
goal(A,B,C):- score_100(C),role(B),my_true_step(A,D),pos_5(D).
********************
********************
New best hypothesis:
tp:26 fn:36 size:15
goal(A,B,C):- score_0(C),role(B),pos_1(D),my_true_step(A,D).
goal(A,B,C):- role(B),score_0(C),pos_4(D),my_true_step(A,D).
goal(A,B,C):- score_100(C),role(B),my_true_step(A,D),pos_5(D).
********************
********************
New best hypothesis:
tp:31 fn:31 size:20
goal(A,B,C):- score_0(C),role(B),pos_1(D),my_true_step(A,D).
goal(A,B,C):- role(B),score_0(C),pos_4(D),my_true_step(A,D).
goal(A,B,C):- score_100(C),role(B),my_true_step(A,D),pos_5(D).
goal(A,B,C):- my_true_step(A,D),score_0(C),role(B),pos_2(D).
********************
********************
New best hypothesis:
tp:62 fn:0 size:25
goal(A,B,C):- pos_3(D),role(B),score_0(C),my_true_step(A,D).
goal(A,B,C):- score_0(C),role(B),pos_1(D),my_true_step(A,D).
goal(A,B,C):- role(B),score_0(C),pos_4(D),my_true_step(A,D).
goal(A,B,C):- score_100(C),role(B),my_true_step(A,D),pos_5(D).
goal(A,B,C):- my_true_step(A,D),score_0(C),role(B),pos_2(D).
********************
Searching programs of size: 6
********************
New best hypothesis:
tp:62 fn:0 size:11
goal(A,B,C):- score_100(C),role(B),my_true_step(A,D),pos_5(D).
goal(A,B,C):- my_true_cell(A,E,D),my_pos(E),c_onecoin(D),score_0(C),role(B).
********************
Searching programs of size: 7
********** SOLUTION **********
Precision:1.00 Recall:1.00 TP:62 FN:0 TN:62 FP:0 Size:11
goal(A,B,C):- score_100(C),role(B),my_true_step(A,D),pos_5(D).
goal(A,B,C):- my_true_cell(A,E,D),my_pos(E),c_onecoin(D),score_0(C),role(B).
******************************
Num. programs: 58378
Constrain:
    Called: 58378 times      Total: 43.46    Mean: 0.001     Max: 8.267
Test:
    Called: 58378 times      Total: 36.91    Mean: 0.001     Max: 0.006
Generate:
    Called: 58379 times      Total: 31.23    Mean: 0.001     Max: 3.489
Combine:
    Called: 16 times     Total: 0.07     Mean: 0.004     Max: 0.009
Total operation time: 111.66s
Total execution time: 120.53s
\end{lstlisting}
\caption{Example \name{} output on the \emph{coins-goal} task.}
\label{fig:coinsg}
\end{figure*}

\begin{figure*}
\begin{lstlisting}
Num. pos examples: 22
Num. neg examples: 22
Searching programs of size: 5
********************
New best hypothesis:
tp:3 fn:19 size:5
goal(A,B,C):- agent_robot(B),prop_4(D),true(A,D),int_0(C).
********************
********************
New best hypothesis:
tp:4 fn:18 size:10
goal(A,B,C):- agent_robot(B),prop_4(D),true(A,D),int_0(C).
goal(A,B,C):- agent_robot(B),true(A,D),int_0(C),prop_3(D).
********************
********************
New best hypothesis:
tp:9 fn:13 size:15
goal(A,B,C):- agent_robot(B),true(A,D),int_0(C),prop_3(D).
goal(A,B,C):- agent_robot(B),prop_5(D),true(A,D),int_0(C).
goal(A,B,C):- agent_robot(B),prop_4(D),true(A,D),int_0(C).
********************
********************
New best hypothesis:
tp:10 fn:12 size:20
goal(A,B,C):- agent_robot(B),true(A,D),int_0(C),prop_3(D).
goal(A,B,C):- agent_robot(B),prop_5(D),true(A,D),int_0(C).
goal(A,B,C):- agent_robot(B),prop_4(D),true(A,D),int_0(C).
goal(A,B,C):- true(A,D),int_0(C),agent_robot(B),prop_1(D).
********************
********************
New best hypothesis:
tp:15 fn:7 size:25
goal(A,B,C):- agent_robot(B),true(A,D),int_0(C),prop_3(D).
goal(A,B,C):- agent_robot(B),prop_5(D),true(A,D),int_0(C).
goal(A,B,C):- true(A,D),int_0(C),agent_robot(B),prop_6(D).
goal(A,B,C):- agent_robot(B),prop_4(D),true(A,D),int_0(C).
goal(A,B,C):- true(A,D),int_0(C),agent_robot(B),prop_1(D).
********************
********************
New best hypothesis:
tp:16 fn:6 size:30
goal(A,B,C):- agent_robot(B),true(A,D),int_0(C),prop_3(D).
goal(A,B,C):- agent_robot(B),prop_5(D),true(A,D),int_0(C).
goal(A,B,C):- int_0(C),prop_2(D),agent_robot(B),true(A,D).
goal(A,B,C):- true(A,D),int_0(C),agent_robot(B),prop_6(D).
goal(A,B,C):- agent_robot(B),prop_4(D),true(A,D),int_0(C).
goal(A,B,C):- true(A,D),int_0(C),agent_robot(B),prop_1(D).
********************
********************
New best hypothesis:
tp:20 fn:2 size:20
goal(A,B,C):- agent_robot(B),prop_5(D),true(A,D),int_0(C).
goal(A,B,C):- int_0(C),not_my_true(A,D),role(B),prop_p(D).
goal(A,B,C):- true(A,D),int_0(C),agent_robot(B),prop_6(D).
goal(A,B,C):- agent_robot(B),prop_4(D),true(A,D),int_0(C).
********************
********************
\end{lstlisting}
\caption{Example \name{} output on the \emph{buttons-goal} task.}
\end{figure*}

\setcounter{figure}{4}
\begin{figure*}
\begin{lstlisting}
New best hypothesis:
tp:21 fn:1 size:20
goal(A,B,C):- not_my_true(A,D),prop_r(D),int_0(C),role(B).
goal(A,B,C):- agent_robot(B),prop_5(D),true(A,D),int_0(C).
goal(A,B,C):- int_0(C),not_my_true(A,D),role(B),prop_p(D).
goal(A,B,C):- true(A,D),int_0(C),agent_robot(B),prop_6(D).
********************
Searching programs of size: 6
Searching programs of size: 7
Searching programs of size: 8
Searching programs of size: 9
********************
New best hypothesis:
tp:22 fn:0 size:24
goal(A,B,C):- not_my_true(A,D),int_0(C),role(B),prop_q(D).
goal(A,B,C):- not_my_true(A,D),prop_r(D),int_0(C),role(B).
goal(A,B,C):- int_0(C),not_my_true(A,D),role(B),prop_p(D).
goal(A,B,C):- agent_robot(B),prop_p(E),true(A,E),true(A,F),prop_r(D),prop_7(F),
              int_100(C),true(A,D).
********************
********** SOLUTION **********
Precision:1.00 Recall:1.00 TP:22 FN:0 TN:22 FP:0 Size:24
goal(A,B,C):- not_my_true(A,D),int_0(C),role(B),prop_q(D).
goal(A,B,C):- not_my_true(A,D),prop_r(D),int_0(C),role(B).
goal(A,B,C):- int_0(C),not_my_true(A,D),role(B),prop_p(D).
goal(A,B,C):- agent_robot(B),prop_p(E),true(A,E),true(A,F),prop_r(D),prop_7(F),
              int_100(C),true(A,D).
******************************
Num. programs: 2094
Test:
    Called: 2094 times   Total: 1.25     Mean: 0.001     Max: 0.009
Constrain:
    Called: 2094 times   Total: 1.09     Mean: 0.001     Max: 0.165
Generate:
    Called: 2095 times   Total: 0.73     Mean: 0.000     Max: 0.035
Combine:
    Called: 11 times     Total: 0.03     Mean: 0.003     Max: 0.005
Total operation time: 3.10s
Total execution time: 3.39s
\end{lstlisting}
\caption{Example \name{} output on the \emph{buttons-goal} task.}
\label{fig:buttons-goa}
\end{figure*}

%% file: manuscript.bbl
\begin{thebibliography}{10}

\bibitem{forest-rubix}
Forest Agostinelli, Rojina Panta, Vedant Khandelwal, Biplav Srivastava,
  Bharath~Chandra Muppasani, Kausik Lakkaraju, and Dezhi Wu, `Explainable
  pathfinding for inscrutable planners with inductive logic programming', in
  {\em ICAPS 2022 Workshop on Explainable AI Planning}, (2022).

\bibitem{atom}
John Ahlgren and Shiu~Yin Yuen, `Efficient program synthesis using constraint
  satisfaction in inductive logic programming', {\em J. Machine Learning Res.},
  {\bf 14}(1),  3649--3682, (2013).

\bibitem{DBLP:conf/ilp/BarthaC19}
S{\'{a}}ndor Bartha and James Cheney, `Towards meta-interpretive learning of
  programming language semantics', in {\em Inductive Logic Programming - 29th
  International Conference, {ILP} 2019, Plovdiv, Bulgaria, September 3-5, 2019,
  Proceedings}, pp. 16--25, (2019).

\bibitem{tilde}
Hendrik Blockeel and Luc {De Raedt}, `Top-down induction of first-order logical
  decision trees', {\em Artif. Intell.}, {\bf 101}(1-2),  285--297, (1998).

\bibitem{zendo}
Neil Bramley, Anselm Rothe, Josh Tenenbaum, Fei Xu, and Todd Gureckis,
  `Grounding compositional hypothesis generation in specific instances', in
  {\em Proceedings of the 40th annual conference of the cognitive science
  society}, (2018).

\bibitem{aspal}
Domenico Corapi, Alessandra Russo, and Emil Lupu, `Inductive logic programming
  in answer set programming', in {\em Inductive Logic Programming - 21st
  International Conference}, pp. 91--97, (2011).

\bibitem{dcc}
Andrew Cropper, `Learning logic programs though divide, constrain, and
  conquer', in {\em Thirty-Sixth {AAAI} Conference on Artificial Intelligence,
  {AAAI} 2022}, pp. 6446--6453. {AAAI} Press, (2022).

\bibitem{ilpintro}
Andrew Cropper and Sebastijan Dumancic, `Inductive logic programming at 30: {A}
  new introduction', {\em J. Artif. Intell. Res.}, {\bf 74},  765--850, (2022).

\bibitem{iggp}
Andrew Cropper, Richard Evans, and Mark Law, `Inductive general game playing',
  {\em Mach. Learn.}, {\bf 109}(7),  1393–1434, (2020).

\bibitem{popper}
Andrew Cropper and Rolf Morel, `Learning programs by learning from failures',
  {\em Mach. Learn.}, {\bf 110}(4),  801--856, (2021).

\bibitem{poppi}
Andrew Cropper and Rolf Morel, `Predicate invention by learning from failures',
  {\em CoRR}, (2021).

\bibitem{reduce}
Andrew Cropper and Sophie Tourret, `Logical reduction of metarules', {\em Mach.
  Learn.}, {\bf 109}(7),  1323--1369, (2020).

\bibitem{meta_abduce}
Wang{-}Zhou Dai and Stephen~H. Muggleton, `Abductive knowledge induction from
  raw data', in {\em {IJCAI} 2021}, (2021).

\bibitem{alps}
Sebastijan Dumančić, Tias Guns, Wannes Meert, and Hendrik Blockeel, `Learning
  relational representations with auto-encoding logic programs', in {\em
  Proceedings of the Twenty-Eighth International Joint Conference on Artificial
  Intelligence, {IJCAI} 2019}, pp. 6081--6087, (2019).

\bibitem{dilp}
Richard Evans and Edward Grefenstette, `Learning explanatory rules from noisy
  data', {\em J. Artif. Intell. Res.},  1--64, (2018).

\bibitem{apperception}
Richard Evans, Jos{\'{e}} Hern{\'{a}}ndez{-}Orallo, Johannes Welbl, Pushmeet
  Kohli, and Marek~J. Sergot, `Making sense of sensory input', {\em Artif.
  Intell.},  103438, (2021).

\bibitem{DBLP:journals/vldb/GalarragaTHS15}
Luis Gal{\'{a}}rraga, Christina Teflioudi, Katja Hose, and Fabian~M. Suchanek,
  `Fast rule mining in ontological knowledge bases with {AMIE+}', {\em {VLDB}
  J.}, {\bf 24}(6),  707--730, (2015).

\bibitem{asp}
Martin Gebser, Roland Kaminski, Benjamin Kaufmann, and Torsten Schaub, {\em
  Answer Set Solving in Practice}, 2012.

\bibitem{clingo}
Martin Gebser, Roland Kaminski, Benjamin Kaufmann, and Torsten Schaub, `Clingo
  = {ASP} + control: Preliminary report', {\em CoRR}, (2014).

\bibitem{ggp}
Michael~R. Genesereth and Yngvi Bj{\"{o}}rnsson, `The international general
  game playing competition', {\em {AI} Magazine}, {\bf 34}(2),  107--111,
  (2013).

\bibitem{DBLP:conf/icml/GlanoisJFWZ0LH22}
Claire Glanois, Zhaohui Jiang, Xuening Feng, Paul Weng, Matthieu Zimmer, Dong
  Li, Wulong Liu, and Jianye Hao, `Neuro-symbolic hierarchical rule induction',
  in {\em International Conference on Machine Learning, {ICML} 2022}, volume
  162, pp. 7583--7615. {PMLR}, (2022).

\bibitem{celine:bottom}
C{\'{e}}line Hocquette and Stephen~H. Muggleton, `Complete bottom-up predicate
  invention in meta-interpretive learning', in {\em Proceedings of the
  Twenty-Ninth International Joint Conference on Artificial Intelligence,
  {IJCAI} 2020}, pp. 2312--2318, (2020).

\bibitem{Kaalia16}
Rama Kaalia, Ashwin Srinivasan, Amit Kumar, and Indira Ghosh, `Ilp-assisted de
  novo drug design', {\em Mach. Learn.}, {\bf 103}(3),  309--341, (2016).

\bibitem{hexmil}
Tobias Kaminski, Thomas Eiter, and Katsumi Inoue, `Meta-interpretive learning
  using hex-programs', in {\em Proceedings of the Twenty-Eighth International
  Joint Conference on Artificial Intelligence, {IJCAI} 2019}, pp. 6186--6190,
  (2019).

\bibitem{michalski:trains}
J.~Larson and Ryszard~S. Michalski, `Inductive inference of {VL} decision
  rules', {\em {SIGART} Newsletter},  38--44, (1977).

\bibitem{ilasp}
Mark Law, Alessandra Russo, and Krysia Broda, `Inductive learning of answer set
  programs', in {\em Logics in Artificial Intelligence - 14th European
  Conference, {JELIA} 2014}, pp. 311--325, (2014).

\bibitem{lloyd:book}
John~W Lloyd, {\em Foundations of logic programming}, Springer Science \&
  Business Media, 2012.

\bibitem{mihalkova2007}
Lilyana Mihalkova, Tuyen Huynh, and Raymond~J Mooney, `Mapping and revising
  markov logic networks for transfer learning', in {\em Aaai}, volume~7, pp.
  608--614, (2007).

\bibitem{mugg:ilp}
Stephen Muggleton, `Inductive logic programming', {\em New Generation
  Computing}, {\bf 8}(4),  295--318, (1991).

\bibitem{progol}
Stephen Muggleton, `Inverse entailment and progol', {\em New Generation
  Comput.}, {\bf 13}(3{\&}4),  245--286, (1995).

\bibitem{ilp20}
Stephen Muggleton, Luc {De Raedt}, David Poole, Ivan Bratko, Peter~A. Flach,
  Katsumi Inoue, and Ashwin Srinivasan, `{ILP} turns 20 - biography and future
  challenges', {\em Mach. Learn.}, {\bf 86}(1),  3--23, (2012).

\bibitem{mugg:metagold}
Stephen~H. Muggleton, Dianhuan Lin, and Alireza Tamaddoni{-}Nezhad,
  `Meta-interpretive learning of higher-order dyadic {Datalog}: predicate
  invention revisited', {\em Mach. Learn.}, {\bf 100}(1),  49--73, (2015).

\bibitem{maxsat}
Nina Narodytska and Fahiem Bacchus, `Maximum satisfiability using core-guided
  maxsat resolution', in {\em Proceedings of the Twenty-Eighth {AAAI}
  Conference on Artificial Intelligence, July 27 -31, 2014, Qu{\'{e}}bec City,
  Qu{\'{e}}bec, Canada}, eds., Carla~E. Brodley and Peter Stone, pp.
  2717--2723. {AAAI} Press, (2014).

\bibitem{louise}
Stassa Patsantzis and Stephen~H. Muggleton, `Top program construction and
  reduction for polynomial time meta-interpretive learning', {\em Mach.
  Learn.}, {\bf 110}(4),  755--778, (2021).

\bibitem{plotkin:thesis}
G.D. Plotkin, {\em Automatic Methods of Inductive Inference}, Ph.D.\
  dissertation, Edinburgh University, August 1971.

\bibitem{hopper}
Stanislaw~J. Purgal, David~M. Cerna, and Cezary Kaliszyk, `Learning
  higher-order logic programs from failures', in {\em Proceedings of the
  Thirty-First International Joint Conference on Artificial Intelligence,
  {IJCAI} 2022, Vienna, Austria, 23-29 July 2022}, ed., Luc~De Raedt, pp.
  2726--2733. ijcai.org, (2022).

\bibitem{prosynth}
Mukund Raghothaman, Jonathan Mendelson, David Zhao, Mayur Naik, and Bernhard
  Scholz, `Provenance-guided synthesis of datalog programs', {\em Proc. {ACM}
  Program. Lang.}, {\bf 4}({POPL}),  62:1--62:27, (2020).

\bibitem{xhail}
Oliver Ray, `Nonmonotonic abductive inductive learning', {\em J. Applied
  Logic}, {\bf 7}(3),  329--340, (2009).

\bibitem{inspire}
Peter Sch{\"{u}}ller and Mishal Benz, `Best-effort inductive logic programming
  via fine-grained cost-based hypothesis generation - the inspire system at the
  inductive logic programming competition', {\em Mach. Learn.}, {\bf 107}(7),
  1141--1169, (2018).

\bibitem{difflog}
Xujie Si, Mukund Raghothaman, Kihong Heo, and Mayur Naik, `Synthesizing datalog
  programs using numerical relaxation', in {\em Proceedings of the
  Twenty-Eighth International Joint Conference on Artificial Intelligence,
  {IJCAI} 2019}, pp. 6117--6124, (2019).

\bibitem{aleph}
Ashwin Srinivasan, `The {ALEPH} manual', {\em Machine Learning at the Computing
  Laboratory, Oxford University}, (2001).

\bibitem{stahl:pi}
Irene Stahl, `The appropriateness of predicate invention as bias shift
  operation in {ILP}', {\em Mach. Learn.}, {\bf 20}(1-2),  95--117, (1995).

\bibitem{tarnlund:hornclause}
Sten{-}{\AA}ke T{\"{a}}rnlund, `Horn clause computability', {\em {BIT}}, {\bf
  17}(2),  215--226, (1977).

\bibitem{quickfoil}
Qiang Zeng, Jignesh~M. Patel, and David Page, `Quickfoil: Scalable inductive
  logic programming', {\em Proc. {VLDB} Endow.}, {\bf 8}(3),  197--208, (2014).

\end{thebibliography}
